\RequirePackage{latexml}
\iflatexml
\documentclass[10pt]{article}
\usepackage{authblk}
\usepackage[authoryear,round]{natbib}
\usepackage{caption}
\usepackage{etoolbox}
\else
\documentclass[10pt]{njulink}

\njulinksetup{
  short-title={Socratic Skill Learning},
  logo={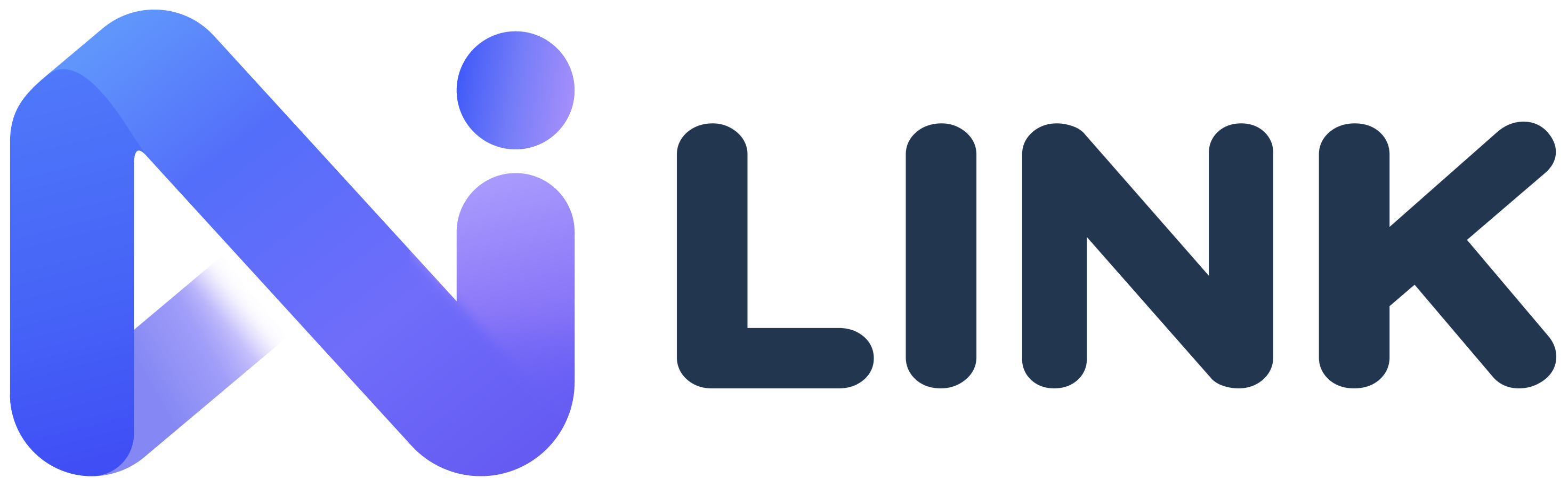}
}
\fi

\usepackage{amsmath,amsfonts,bm}

\def\eqref#1{equation~\ref{#1}}

\def\1{\bm{1}}

\DeclareMathAlphabet{\mathsfit}{\encodingdefault}{\sfdefault}{m}{sl}
\SetMathAlphabet{\mathsfit}{bold}{\encodingdefault}{\sfdefault}{bx}{n}

\usepackage{amsmath,amssymb}
\usepackage{booktabs}
\usepackage{colortbl}
\usepackage{siunitx}
\usepackage{graphicx}
\usepackage{microtype}
\usepackage{xcolor}
\usepackage{xspace}
\usepackage{enumitem}
\usepackage{array}
\usepackage{multirow}
\usepackage{float}
\usepackage{wrapfig}
\usepackage{needspace}
\pretocmd{\subsection}{\Needspace{5\baselineskip}}{}{}
\usepackage{placeins}

\usepackage{tabularx}
\usepackage{longtable}
\usepackage{pgfplots}
\pgfplotsset{compat=1.18}
\usepackage{tikz}
\usetikzlibrary{arrows.meta,calc,fit,positioning,shapes.geometric}
\usepackage{hyperref}
\usepackage{xurl}
\newcommand{\SSLPDFAuthors}{Letian Zhu, Han Li, Yifan Yao, Xinping Lei,
  Xinyu Che, Jiahao Wang, Dianyi Wang, Yucheng Fu, Jiakai Wang,
  Zhiqi Bai, Wenbo Su, Jiaheng Liu}
\newcommand{\SSLTitle}{Socratic Skill Learning: Interleaving Teaching and Practice for Transferable Agent Skills}
\iflatexml
\newcommand{\email}[1]{\href{mailto:#1}{\nolinkurl{#1}}}
\author[1]{Letian~Zhu}
\author[1]{Han~Li}
\author[1]{Yifan~Yao}
\author{Xinping~Lei}
\author[1]{Xinyu~Che}
\author[1]{Jiahao~Wang}
\author[1]{Dianyi~Wang}
\author[1]{Yucheng~Fu}
\author[2]{Jiakai~Wang}
\author[2]{Zhiqi~Bai}
\author[2]{Wenbo~Su}
\author[1]{Jiaheng~Liu}
\affil[1]{Nanjing University}
\affil[2]{Alibaba Group}
\else
\author[1,*]{Letian~Zhu}
\author[1,*]{Han~Li}
\author[1,*]{Yifan~Yao}
\author{Xinping~Lei}
\author[1]{Xinyu~Che}
\author[1]{Jiahao~Wang}
\author[1]{Dianyi~Wang}
\author[1]{Yucheng~Fu}
\author[2]{Jiakai~Wang}
\author[2]{Zhiqi~Bai}
\author[2]{Wenbo~Su}
\author[1,\dagger]{Jiaheng~Liu}
\affiliation[1]{Nanjing University}
\affiliation[2]{Alibaba Group}
\contribution[*]{Equal contribution.}
\contribution[\dagger]{Corresponding author.}
\correspondence{Jiaheng Liu at \email{liujiaheng@nju.edu.cn}}
\fi

\newcommand{\SSLRepositoryURL}{https://github.com/NJU-LINK/Socratic-Skill-Learning}
\newcommand{\SSLRepositoryLink}{\expandafter\url\expandafter{\SSLRepositoryURL}}

\AtBeginDocument{\hypersetup{
  pdftitle={\SSLTitle},pdfauthor={\SSLPDFAuthors}}}

\newcommand{\tableappearance}{%
  \color{black}\arrayrulecolor{black}%
  \hypersetup{allcolors=black}}
\newcommand{\tablecaptionspacing}{\captionsetup{skip=7pt}}
\AddToHook{env/table/begin}{\tableappearance}
\AddToHook{env/table*/begin}{\tableappearance}
\AddToHook{env/longtable/begin}{\tableappearance}

\definecolor{pilotblue}{RGB}{32,88,140}
\definecolor{oursred}{HTML}{B26649}
\definecolor{teacherorange}{HTML}{AA5F47}
\definecolor{studentblue}{HTML}{507294}
\definecolor{skillgreen}{RGB}{79,139,89}
\definecolor{softgray}{RGB}{241,242,244}
\definecolor{textgray}{RGB}{80,84,90}
\definecolor{treegreen}{HTML}{62724D}

\newcommand{\developalong}{\textsc{SSL}\xspace}
\newcommand{\skilltransferbench}{\textsc{SkillTransferBench}\xspace}
\newcommand{\fixedwindow}{\textsc{Fixed-Window SSL}\xspace}
\newcommand{\issuetree}{\textsc{Learning-Issue Tree}\xspace}
\newcommand{\treedevelop}{\textsc{Issue-Guided SSL}\xspace}
\newcommand{\compileafter}{\textsc{Expert Authoring}\xspace}
\newcommand{\teachafter}{\textsc{Post-hoc Teaching}\xspace}
\newcommand{\tracebaseline}{\textsc{Trace2Skill}\xspace}
\newcommand{\skillkdbaseline}{\textsc{SKILL-KD}\xspace}
\newcommand{\GFiveUtility}{0.776}

\newcommand{\GFiveTokens}{9.56}

\newcommand{\GFiveObservations}{1126}
\newcommand{\GFiveLessons}{403}
\newcommand{\GTenUtility}{0.801}

\newcommand{\GTenTokens}{6.89}

\newcommand{\GTenObservations}{630}
\newcommand{\GTenLessons}{254}
\newcommand{\GTwentyUtility}{0.776}

\newcommand{\GTwentyTokens}{6.61}

\newcommand{\GTwentyObservations}{375}
\newcommand{\GTwentyLessons}{169}
\newcommand{\TreeUtility}{0.805}

\newcommand{\TreeTokens}{5.72}

\newcommand{\TreeObservations}{329}
\newcommand{\TreeLessons}{135}
\newcommand{\TreeEducatorEfficiencyRatio}{2.75}

\newcommand{\BaseUtility}{0.723}

\newcommand{\CompileUtility}{0.728}

\newcommand{\CompileGain}{0.46}
\newcommand{\CompileCI}{[-5.46,6.56]}

\newcommand{\TeachUtility}{0.748}

\newcommand{\TeachGain}{2.44}
\newcommand{\TeachCI}{[-3.36,7.90]}

\newcommand{\TraceUtility}{0.761}

\newcommand{\TraceGain}{3.76}
\newcommand{\TraceCI}{[-0.50,8.96]}

\newcommand{\DevelopUtility}{0.801}

\newcommand{\DevelopGain}{7.75}
\newcommand{\DevelopCI}{[3.53,12.68]}

\newcommand{\IssueTreeUtility}{0.805}

\newcommand{\IssueTreeGain}{8.18}
\newcommand{\IssueTreeCI}{[3.54,13.57]}

\newcommand{\SkillKDUtility}{0.751}
\newcommand{\SkillKDGain}{2.76}
\newcommand{\SkillKDCI}{[-1.59,8.09]}

\newcommand{\BaseFullPass}{32.50}
\newcommand{\CompileFullPass}{40.44}
\newcommand{\TeachFullPass}{40.37}
\newcommand{\TraceFullPass}{39.11}
\newcommand{\DevelopFullPass}{49.49}
\newcommand{\IssueTreeFullPass}{50.82}
\newcommand{\SkillKDFullPass}{41.65}
\newcommand{\DevelopFullPassGain}{16.99}
\newcommand{\DevelopFullPassCI}{[8.13,26.86]}
\newcommand{\IssueTreeFullPassGain}{18.33}
\newcommand{\IssueTreeFullPassCI}{[9.44,28.12]}
\newcommand{\DevelopOverCompile}{7.29}
\newcommand{\DevelopOverCompileCI}{[3.62,11.44]}
\newcommand{\DevelopOverTeach}{5.31}
\newcommand{\DevelopOverTeachCI}{[2.34,9.09]}
\newcommand{\DevelopOverTrace}{3.99}
\newcommand{\DevelopOverTraceCI}{[1.45,6.49]}
\newcommand{\DevelopOverSkillKD}{4.99}
\newcommand{\DevelopOverSkillKDCI}{[1.79,8.48]}

\newcommand{\TeachOverCompileCI}{[-1.57,5.64]}
\newcommand{\HeadroomCases}{130}
\newcommand{\CeilingCases}{25}
\newcommand{\DevelopHeadroomPercent}{55.4}
\newcommand{\DevelopHeadroomRow}{72 & 37 & 21}
\newcommand{\DevelopCeilingRow}{0 & 22 & 3}
\newcommand{\IssueTreeHeadroomPercent}{54.6}
\newcommand{\IssueTreeHeadroomRow}{71 & 43 & 16}
\newcommand{\IssueTreeCeilingRow}{0 & 20 & 5}
\newcommand{\DevelopPositiveFamilies}{16}
\newcommand{\CatalogFamilies}{35}
\newcommand{\CatalogCases}{408}
\newcommand{\MainFamilies}{22}
\newcommand{\MainVariants}{155}
\newcommand{\MainRepeats}{3}
\newcommand{\CatalogNonAnchorCases}{373}
\newcommand{\CatalogCandidateFamilies}{37}
\newcommand{\CatalogCandidateCases}{444}

\newcommand{\CatalogReferenceChecks}{2,807}
\newcommand{\CatalogChecksMin}{3}
\newcommand{\CatalogChecksMax}{34}
\newcommand{\CatalogCasesMin}{5}
\newcommand{\CatalogCasesMax}{17}
\newcommand{\MainHeldoutMin}{5}
\newcommand{\MainHeldoutMax}{10}
\newcommand{\MainRuns}{66}
\newcommand{\MainRounds}{3}
\newcommand{\MainAuthoringRuns}{198}
\newcommand{\ISTEducatorTotalM}{158.2}
\newcommand{\TreeEducatorTotalM}{60.6}
\newcommand{\ReviewReduction}{47.8}
\newcommand{\ISTEducatorPerReviewK}{251.1}
\newcommand{\TreeEducatorPerReviewK}{184.1}
\newcommand{\ReviewTokenReduction}{26.7}
\newcommand{\TreeEducatorReduction}{61.7}
\newcommand{\TreeTotalCostReduction}{16.9}
\newcommand{\ISTPosthocCostReduction}{21}

\newcommand{\DeepSeekISTGain}{3.08}
\newcommand{\KimiISTGain}{8.24}
\newcommand{\CrossFamilies}{10}
\newcommand{\CrossVariants}{77}
\newcommand{\CrossRepeats}{1}

\newcommand{\CrossIterations}{3}
\newcommand{\CrossAcquisitionRuns}{40}
\newcommand{\CrossAnchorEvaluations}{120}
\newcommand{\CrossSkillEvaluations}{308}
\newcommand{\CrossNoSkillEvaluations}{77}
\newcommand{\CrossCohortSizes}{Audio STFT peaks (10 cases), Filter JS from HTML (10), Git leak recovery (6), HTML finance verification (6), JQ processing (7), KV-store gRPC (6), PyPI server (10), SPARQL professors (10), SQLite truncation (6), and XLSX structure (6)}

\newcommand{\SubsetDevelopOverTeach}{4.88}

\newcommand{\DeepSeekDevelopOverTeach}{2.54}
\newcommand{\DeepSeekEducatorReduction}{63.8}
\newcommand{\DeepSeekStudentIncrease}{39.2}

\newcommand{\DeepSeekTeachingEfficiencyRatio}{2.2}

\newcommand{\KimiDevelopOverTeach}{4.09}
\newcommand{\KimiEducatorReduction}{77.4}
\newcommand{\KimiStudentReduction}{16.1}

\newcommand{\KimiTeachingEfficiencyRatio}{3.9}
\newcommand{\DeepSeekTreeReviews}{69}

\newcommand{\DeepSeekISTWakeups}{195}
\newcommand{\DSStudentCallsIST}{890}
\newcommand{\DSStudentCallsTree}{1,216}
\newcommand{\DSStudentCallIncreasePercent}{36.6}
\newcommand{\DSStudentAddedM}{52.8}
\newcommand{\DSEducatorSavedM}{41.8}
\newcommand{\DSNetAddedM}{11.0}
\newcommand{\DSSqliteAddedM}{23.8}
\newcommand{\DSOtherFamiliesSavedM}{12.8}
\newcommand{\DSGiantCalls}{25}
\newcommand{\DSGiantCallMaxK}{799}
\newcommand{\DSGiantCallTokenShare}{35}

\newcommand{\AnchorRoundOne}{0.824}
\newcommand{\AnchorRoundTwo}{0.986}
\newcommand{\AnchorRoundThree}{0.990}
\newcommand{\CompileTokens}{4.95}
\newcommand{\TeachTokens}{8.71}
\newcommand{\SkillKDRecordedAcquisitionM}{128.11}
\newcommand{\SkillKDRecordedTokensPerRunM}{1.94}
\newcommand{\CohortDirectionRows}{%
\compileafter & 58 & 56 & 41 \\
\teachafter & 57 & 53 & 45 \\
Adapted \tracebaseline & 59 & 65 & 31 \\
\fixedwindow & 72 & 59 & 24 \\
\treedevelop & 71 & 63 & 21 \\%
}
\newcommand{\SkillKDDirectionRow}{Adapted \skillkdbaseline & 55 & 69 & 31 \\}
\newcommand{\CohortRepeatRows}{%
No Skill & 0.719 & 0.730 & 0.721 & 0.006 & 0.036 \\
\fixedwindow & 0.772 & 0.799 & 0.831 & 0.030 & 0.031 \\
\treedevelop & 0.767 & 0.810 & 0.838 & 0.036 & 0.037 \\%
}
\newcommand{\SkillKDRepeatRow}{Adapted \skillkdbaseline & 0.746 & 0.756 & 0.750 & 0.005 & 0.039 \\}
\newcommand{\ComparisonFamilyRows}{%
ACL inheritance & 0.933 & 0.728 & 0.858 & 0.885 & 0.933 & 0.933 & 0.881 \\
API scheduler & 0.767 & 0.730 & 0.745 & 0.897 & 0.921 & 0.791 & 0.830 \\
Audio STFT peaks & 0.563 & 0.631 & 0.652 & 0.608 & 0.664 & 0.643 & 0.656 \\
\shortstack[l]{Bank transaction\\filter} & 0.678 & 0.548 & 0.550 & 0.522 & 0.515 & 0.593 & 0.652 \\
\shortstack[l]{Cross-entropy\\method} & 0.553 & 0.556 & 0.642 & 0.617 & 0.556 & 0.667 & 0.656 \\
DB WAL recovery & 0.473 & 0.826 & 0.705 & 0.905 & 0.901 & 0.884 & 0.878 \\
\shortstack[l]{Deterministic\\tarball} & 0.678 & 0.691 & 0.677 & 0.706 & 0.682 & 0.728 & 0.693 \\
\shortstack[l]{Filter JS\\from HTML} & 0.366 & 0.350 & 0.403 & 0.363 & 0.391 & 0.433 & 0.440 \\
\shortstack[l]{Financial\\analysis} & 0.792 & 0.750 & 0.875 & 0.747 & 0.792 & 0.875 & 0.875 \\
\shortstack[l]{Financial\\documents} & 0.921 & 0.603 & 0.754 & 0.913 & 0.897 & 0.913 & 0.873 \\
Flask & 0.408 & 0.406 & 0.638 & 0.516 & 0.396 & 0.690 & 0.785 \\
Git leak recovery & 0.929 & 0.905 & 0.872 & 0.896 & 0.928 & 0.975 & 0.961 \\
\shortstack[l]{HTML finance\\verification} & 0.731 & 1.000 & 0.944 & 0.935 & 0.824 & 0.917 & 0.944 \\
JQ processing & 0.875 & 0.835 & 0.853 & 0.889 & 0.867 & 0.889 & 0.892 \\
KV-store gRPC & 0.607 & 0.635 & 0.630 & 0.586 & 0.584 & 0.625 & 0.635 \\
ODE RK4 & 0.887 & 0.987 & 1.000 & 0.973 & 0.827 & 1.000 & 1.000 \\
Office workflow & 0.875 & 0.905 & 0.907 & 0.867 & 0.873 & 0.959 & 0.964 \\
Organization JSON & 0.778 & 1.000 & 1.000 & 0.875 & 1.000 & 1.000 & 1.000 \\
PyPI server & 0.544 & 0.578 & 0.564 & 0.536 & 0.533 & 0.583 & 0.589 \\
SPARQL professors & 0.706 & 0.680 & 0.690 & 0.713 & 0.672 & 0.680 & 0.705 \\
SQLite truncation & 0.915 & 0.747 & 0.569 & 0.856 & 0.830 & 0.915 & 0.875 \\
XLSX structure & 0.933 & 0.923 & 0.919 & 0.933 & 0.933 & 0.923 & 0.928 \\%
}
\newcommand{\ComparisonFamilyMacro}{%
Macro average & 0.723 & 0.728 & 0.748 & 0.761 & 0.751 & 0.801 & \textbf{0.805} \\
}
\newcommand{\ReviewIntentRows}{%
Knowledge transfer & Facts, mechanisms, APIs, formats, boundaries, or expert experience & 101 (39.8\%) & 19 (20.9\%) \\
Reasoning guidance & Representation, decomposition, causal reasoning, comparison, generalization, or decisions & 168 (66.1\%) & 43 (47.3\%) \\
Exploration/verification & Search, experimental design, evidence collection, validation, or tool-use procedure & 23 (9.1\%) & 21 (23.1\%) \\
Error diagnosis/recovery & Attribution of an observed error or stuck state and a recovery direction & 99 (39.0\%) & 46 (50.5\%) \\
Epistemic calibration & Unsupported assumptions, over-generalization, evidence levels, or confidence & 39 (15.4\%) & 11 (12.1\%) \\
Experience consolidation & Retrievable triggers, rules, workflows, scripts, or references from local experience & 104 (40.9\%) & 25 (27.5\%) \\
Other & A target not captured by a combination of the six core labels & 0 & 0 \\%
}
\newcommand{\CaseCorrectLine}{210}
\newcommand{\CaseInitialFlags}{4}

\newcommand{\CaseFreshPassed}{4}
\newcommand{\CaseFreshTotal}{4}

\newcommand{\FlaskLocalChecks}{14}

\newcommand{\FlaskAnchorOne}{0.667}
\newcommand{\FlaskAnchorTwo}{1.000}
\newcommand{\FlaskAnchorThree}{1.000}
\newcommand{\IssueReopenEvents}{28}
\newcommand{\IssueBypassClosures}{9}

\newcommand{\IssueLifecycleNodeRows}{%
Persistent issue nodes & 177 \\
Issues/run: mean; median; range & 2.68; 3; 1--6 \\
Nodes reopened at least once & 23/177 (13.0\%) \\
Reopen events & 28 \\
Reopen events followed by a new lesson & 28/28 \\
Nodes closed by run end & 175 \\
Nodes still pending at run end & 2 \\%
}
\newcommand{\IssueLifecycleClosureRows}{%
Educator silence & 194 \\
Verification-backed closures & 7 \\
Bounded-scope closures & 2 \\
Evidence-backed rejection closures & 0 \\%
}

\newcommand{\ReviewRuns}{66}
\newcommand{\ReviewFamilies}{22}
\newcommand{\ReviewTotal}{630}
\newcommand{\ReviewOpeningLessons}{254}
\newcommand{\ReviewFollowUps}{91}
\newcommand{\ReviewIntervalCount}{432}
\newcommand{\ReviewTerminalCount}{198}
\newcommand{\ReviewTerminalTeaching}{28}
\newcommand{\ReviewTerminalSilence}{170}
\newcommand{\RoundOneRegularTeach}{170}
\newcommand{\RoundOneRegularReviews}{233}
\newcommand{\RoundTwoRegularTeach}{42}
\newcommand{\RoundTwoRegularReviews}{112}
\newcommand{\RoundThreeRegularTeach}{14}
\newcommand{\RoundThreeRegularReviews}{87}
\newcommand{\RoundOneTokenM}{22.3}
\newcommand{\RoundTwoTokenM}{49.9}
\newcommand{\RoundThreeTokenM}{60.5}
\newcommand{\RoundOneSilentShare}{40.4}
\newcommand{\RoundTwoSilentShare}{69.8}
\newcommand{\RoundThreeSilentShare}{85.8}
\newcommand{\ReviewTeachIntervals}{226}
\newcommand{\ReviewSilentIntervals}{206}
\newcommand{\ReviewFirstRoundRate}{73.0}
\newcommand{\ReviewLastRoundRate}{16.1}

\newcommand{\ReviewWriteAdjustedPP}{+23.9}

\newcommand{\ReviewPackageEligible}{193}
\newcommand{\ReviewPackageKnown}{112}
\newcommand{\ReviewPackageTeachN}{69}
\newcommand{\ReviewPackageSilentN}{43}
\newcommand{\ReviewPackageTeachChanged}{67}
\newcommand{\ReviewPackageSilentChanged}{27}

\newcommand{\ReviewPackageAdjustedPP}{+38.6}
\newcommand{\ReviewFirstCostK}{74.7}
\newcommand{\ReviewLastCostK}{395.7}
\newcommand{\ReviewCostGrowth}{5.3}
\newcommand{\ReviewLastSilentTokenShare}{85.8}

\newcommand{\ReviewWithinRunPackagePP}{34.3}
\newcommand{\ReviewWithinRunPackageN}{18}
\newcommand{\ReviewToolErrorWindows}{111}
\newcommand{\ReviewNoErrorWindows}{321}
\newcommand{\ReviewTerminalTeachingPreceded}{24}
\newcommand{\ReviewOtherWindows}{81}

\newcommand{\ReviewAllIntervalKnown}{316}
\newcommand{\ReviewTransitions}{132}
\newcommand{\StageOneTeach}{188}
\newcommand{\StageTwoTeach}{47}
\newcommand{\StageThreeTeach}{19}
\newcommand{\StageOneReviews}{299}
\newcommand{\StageTwoReviews}{178}
\newcommand{\StageThreeReviews}{153}
\newcommand{\StageOnePerfect}{38}
\newcommand{\StageTwoPerfect}{61}
\newcommand{\StageThreePerfect}{63}
\newcommand{\StageFirstLessons}{188}
\newcommand{\StageLaterLessons}{66}
\newcommand{\StageEarlyReasoningCount}{140}
\newcommand{\StageLateReasoningCount}{28}
\newcommand{\StageEarlyDiagnosisCount}{62}
\newcommand{\StageLateDiagnosisCount}{37}
\newcommand{\PostPerfectTransitions}{99}
\newcommand{\PostPerfectSecond}{38}
\newcommand{\PostPerfectThird}{61}
\newcommand{\PostPerfectRuns}{61}
\newcommand{\PostPerfectLessonChanged}{32}
\newcommand{\PostPerfectLessonTotal}{33}
\newcommand{\PostPerfectSilentChanged}{66}
\newcommand{\PostPerfectSilentTotal}{66}
\newcommand{\PostPerfectLessons}{35}
\newcommand{\PostPerfectLaterSilent}{76}
\newcommand{\ReviewSilentOpportunityShare}{59.7}
\newcommand{\ReviewSilentOpeningShare}{72.1}

\newcommand{\ProgressRevisedAfterPass}{98}

\newcommand{\ReviewWriteCI}{[13.0,36.1]}
\newcommand{\ReviewPackageCI}{[10.7,57.8]}

\newcommand{\skillpkg}{\mathcal{K}}

\title{\SSLTitle}
\date{October 12, 2026}
\iflatexml\else
\metadata[Code]{\SSLRepositoryLink}
\fi

\newcommand{\SSLAbstractText}{%
Agent Skills improve LLM agents by making task knowledge reusable across
sessions. Learning transferable Skills requires evidence about a procedure
and its conditions of use, which a successful task attempt can leave
untested. We introduce Socratic Skill Learning (\developalong), which
interleaves expert teaching with Student practice. An Educator guides
investigation while the Student remains the sole environment operator and
Skill author, able to question, test, and adapt advice. To test reuse beyond
the learning task, we construct \skilltransferbench{}, with \CatalogFamilies{} task
families and \CatalogCases{} executable cases, pairing public variation
specifications with held-out evaluation in fresh solver sessions. On the
main \MainFamilies{}-family, \MainVariants{}-case cohort, Fixed-Window SSL
reaches \DevelopUtility{} utility, versus \BaseUtility{} without a Skill,
\TraceUtility{} for adapted Trace2Skill (trajectory distillation), and
\SkillKDUtility{} for adapted SKILL-KD (contrastive distillation).
Across three model pairs on shared families, interleaved guidance outperforms
expert authoring and post-hoc teaching. Process analysis links guidance to
Student writing and package revision; a case study shows the Student
testing and refining a proposed procedure for later reuse. As demand
for new lessons falls, the cost of
fixed-window review grows. Issue-Guided SSL tracks unresolved questions
across Skill revisions to select and focus review,
reaching \IssueTreeUtility{} utility with \TreeEducatorReduction{}\%
fewer Educator tokens and \TreeTotalCostReduction{}\% fewer total
acquisition tokens than Fixed-Window SSL on the main cohort.
}

\iflatexml\else
\abstract{\SSLAbstractText}
\fi

\begin{document}
\raggedbottom

\maketitle
\iflatexml
\begin{center}
Equal contribution: Letian Zhu, Han Li, and Yifan Yao.\par
Corresponding author: Jiaheng Liu at \email{liujiaheng@nju.edu.cn}\par
Code: \SSLRepositoryLink
\end{center}
\begin{abstract}
\SSLAbstractText
\end{abstract}
\fi

\newlength{\papertextfloatsep}\setlength{\papertextfloatsep}{\textfloatsep}
\newlength{\paperfloatsep}\setlength{\paperfloatsep}{\floatsep}
\newlength{\paperintextsep}\setlength{\paperintextsep}{\intextsep}
\newlength{\papercaptionskip}\setlength{\papercaptionskip}{\abovecaptionskip}
\setlength{\floatsep}{\paperfloatsep}
\setlength{\abovecaptionskip}{\papercaptionskip}

\section{Introduction}
\label{sec:introduction}

Agent Skills extend LLM agents through reusable
instructions and executable resources, without updating model weights
\citep{anthropic2025skills}. Curated Skills improve
success on tasks requiring specialized expertise
\citep{skillsbench2026}. Acquiring transferable Skills requires
identifying which steps to reuse,
under what conditions, and when to adapt them.

Recent methods distill execution trajectories into Skills and refine
packages through testing and feedback
\citep{trace2skill2026,skillgen2026,skillkd2026}. SkillLearnBench's Teacher
Feedback uses expert advice after failed attempts to guide
revision \citep{skilllearnbench2026}. Such refinement depends on
which hypotheses the learner tests and
which failures it investigates. Guidance during an ongoing attempt can
redirect these choices before an untested assumption becomes a reusable
rule. We therefore ask: \emph{how should external expertise guide the
learner while it is still gathering evidence?}

We introduce \textbf{Socratic Skill Learning (\developalong)}, interleaving
expert teaching with Student practice (Figure~\ref{fig:insight}).\footnote{Code,
benchmark, and trajectories:
\href{\SSLRepositoryURL}{project repository}.}
An Educator observes recent Student work and the live Skill, then explains,
challenges, diagnoses, or remains silent. The Student operates the environment
and authors the package. Inspired by Socratic inquiry, the interaction
examines assumptions through dialogue and practical checks; the Student
can question advice, adapt it, and decide what to retain.
We instantiate SSL with \fixedwindow, which reviews at regular tool-call
intervals, and \treedevelop, which focuses review on unresolved learning issues.

\suppressfloats[t]
\begin{figure}[!htbp]
\centering
\includegraphics[width=\linewidth]{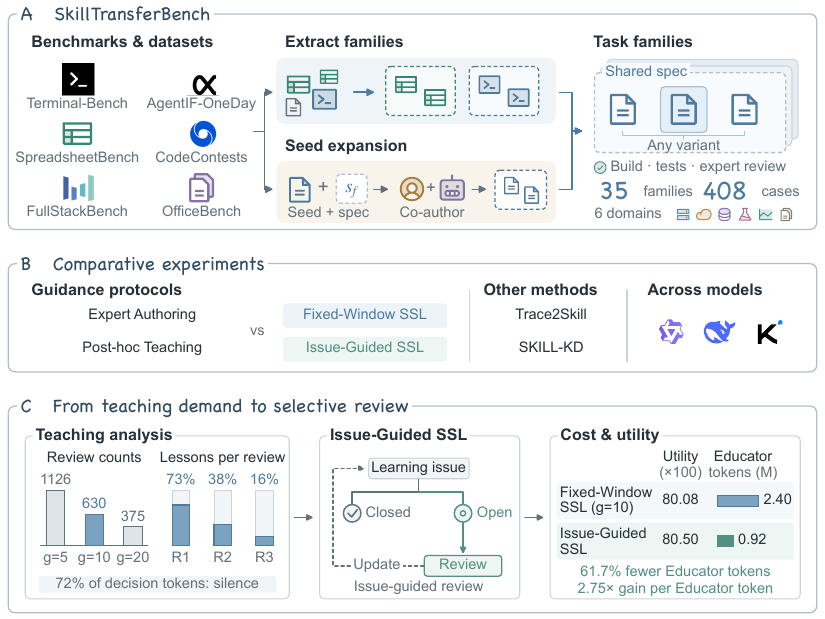}
\caption{\textbf{Overview of the study.}
Panels show (A) task-family construction from six source benchmarks, (B)
acquisition-protocol comparisons, external methods, and model pairs, and
(C) the path from teaching demand to issue-guided review with utility and cost.
Panel C summarizes Qwen experiments on the \MainFamilies{}-family main cohort.}
\label{fig:overview}
\end{figure}

Existing benchmarks measure Skill utility and reuse
\citep{skillsbench2026,skilllearnbench2026}. Studying guided acquisition
also requires specifying what the learner should investigate beyond the
practice task. We construct
\skilltransferbench{}, a benchmark of \CatalogFamilies{} \textbf{task families} and
\CatalogCases{} executable cases across six domains. Each family pairs an
anchor task with a public specification of invariants and permitted variation.
This specification guides exploration; fresh solvers test the acquired
package on held-out family members.

\Needspace{5\baselineskip}
We organize the study around four questions (Figure~\ref{fig:overview}):
\textbf{RQ1:} How well does SSL transfer relative to existing induction methods?
\textbf{RQ2:} How should external expertise guide Skill acquisition?
\textbf{RQ3:} How does guidance enter Skill development as learning progresses?
\textbf{RQ4:} How should review be organized to sustain transfer with less
expert supervision?

On shared families across three model pairs, both interleaved protocols
outperform direct expert authoring and post-hoc teaching. Process analysis
links lessons to Student development; a case study shows the Student
testing an Educator-proposed procedure and refining it for later reuse.
\treedevelop{} reaches \IssueTreeUtility{} utility with
\TreeEducatorReduction{}\% fewer Educator tokens on the main cohort.

\Needspace{16\baselineskip}
\noindent\textbf{Our contributions are threefold:}
\begin{itemize}[leftmargin=*,topsep=3pt,itemsep=2pt,parsep=0pt]
\item We construct \skilltransferbench{}, containing \CatalogFamilies{}
task families and \CatalogCases{} executable cases across six domains,
with public family specifications that guide acquisition and held-out
evaluation that tests frozen Skills in fresh solver sessions.
\item We introduce Socratic Skill Learning (\developalong), in which the
Student owns exploration, verification, and Skill authoring while an
Educator guides practice. Issue-Guided SSL uses persistent learning issues
to select and focus review around unresolved questions.
\item We evaluate Skill transfer across three model pairs and complement
these comparisons with process analysis and case studies, linking guidance
to Student development and showing how selecting and focusing review
sustains transfer with less expert supervision.
\end{itemize}

\FloatBarrier
\Needspace{7\baselineskip}
\section{Related Work}
\label{sec:background}

\paragraph{Constructing Skills from execution evidence.}
Skills can retain experience as guidance or executable code
\citep{zhao2024expel,wang2023voyager}, consolidate trajectory lessons
\citep{trace2skill2026}, or evolve through execution and verification
\citep{skillgen2026,skillrevise2026}. SKILL-KD's consolidation agent
derives patches from Teacher--Student trajectory contrasts and validates
them through Student reruns \citep{skillkd2026}.

\paragraph{Feedback during Skill development.}
Self-Refine and Reflexion improve outputs or attempts through feedback
\citep{madaan2023selfrefine,shinn2023reflexion}. For agent diagnosis \citep{zhao2026failureprocess},
CodeTracer reconstructs state transitions to localize failures
\citep{li2026codetracer}. SkillLearnBench uses a
reference Skill to guide revision after failure \citep{skilllearnbench2026};
AgentBrew validates teacher-authored corrections through student retries
\citep{agentbrew2026}. HASP includes teacher review of student-proposed
skill programs \citep{hasp2026}, while SkillHone preserves revision
evidence and decisions \citep{skillhone2026}. SSL places Educator guidance
within Student investigation and verification, leaving the Student to
author the Skill. Unresolved learning questions select and focus review.
This design connects Socratic dialogue \citep{socraticlm2024} with learner
construction \citep{chi2001learning} and feedback on the learning process
\citep{hattie2007feedback}.

\paragraph{Scope and transfer in Skill evaluation.}
SkillsBench measures Skill utility \citep{skillsbench2026}; SkillLearnBench
and EconSkills evaluate reuse across parameterized instances
\citep{skilllearnbench2026,econskills2026}. Trace2Skill tests held-out
transfer \citep{trace2skill2026}, while ContinualSkillBench uses ordered
task streams \citep{guan2026continualskillbench}. \skilltransferbench{}
defines acquisition scope through a public family specification. Across
six domains, evaluation in fresh sessions connects guided acquisition to
held-out transfer and supervision cost.

\section{SkillTransferBench}
\label{sec:benchmark}

\subsection{Task families and Skill transfer}
\label{sec:setup}

Skill acquisition develops a reusable procedure for a task family $f$.
Acquisition uses an executable anchor $x_{f,0}$ for practice and a public
specification $s_f$ defining invariant goals and permitted variation
(e.g., WAL encoding, integrity, and recovery depth). The specification directs
investigation beyond the anchor; held-out family members test transfer
across these variations.

The acquired Skill is a package $\skillpkg=(M,\mathcal R)$, containing
instructions $M=\texttt{SKILL.md}$ and optional scripts, references, and
assets $\mathcal R$. We evaluate the frozen package with the target solver $\pi_S$
in a new session for each case (a fresh solver). Transfer utility is
\begin{equation}
 U_f(\skillpkg;\pi_S)
 = \mathbb E_{x\sim\mathcal D_f}\,[r_f(\pi_S(x,\skillpkg))],
\end{equation}
where $\mathcal D_f$ is the case distribution and $r_f$ the verifier score.

\subsection{Benchmark construction}

We construct the benchmark in three stages: adapt source tasks, build task
families through two complementary routes, and validate the resulting cases.

\paragraph{Source adaptation.}
We select tasks with clear requirements and meaningful variation from six
benchmarks: Terminal-Bench and AgentIF-OneDay
\citep{terminalbench2026,agentifoneday2026}, SpreadsheetBench
\citep{spreadsheetbench2024}, CodeContests and FullStackBench
\citep{codecontests2022,fullstackbench2024}, and OfficeBench
\citep{officebench2024}. We normalize their formats and align descriptions,
fixtures, and scorers.

\paragraph{Task-family construction.}
Experts in coding and agent benchmarking collaborate with agents through two
routes (Figure~\ref{fig:overview}A): \textbf{extraction and adaptation}
groups compatible source cases under a shared specification and aligns
their statements and tests; \textbf{seed expansion} starts from one task,
defines its invariant goal and permitted variation, and authors additional
cases within that scope. Both routes yield a public family specification
and executable variants. One fixed case serves as the acquisition anchor;
the remaining cases provide candidates for held-out evaluation.

\paragraph{Validation.}
Each case includes a resettable environment, a reference solution, and
private tests. Programmatic checks verify Docker builds, execute tests,
and confirm that reference solutions pass them; an expert other than the
author then reviews each case before acceptance. Reference solutions and
private tests remain outside agent workspaces (Appendix~\ref{app:assets}).

\begin{figure}[!t]
\centering
\newcommand{\benchsidetitle}[1]{%
  \makebox[\linewidth][l]{\hspace*{0.186667\linewidth}\textbf{#1}}}
\begin{minipage}[t]{0.575\linewidth}\vspace{0pt}
\centering
\includegraphics[width=\linewidth]{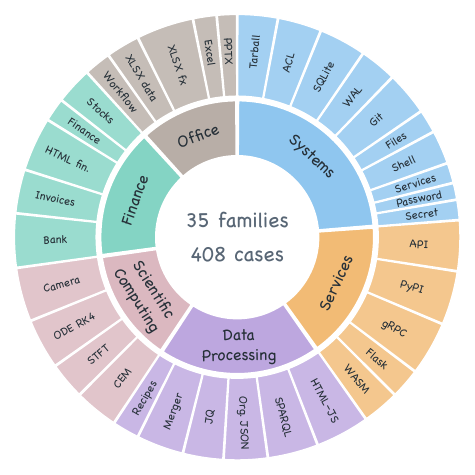}\par
\smallskip
\small\textbf{(a) Domain and family composition}
\end{minipage}\hfill
\begin{minipage}[t]{0.409\linewidth}\vspace{0pt}
\centering
\includegraphics[width=\linewidth]{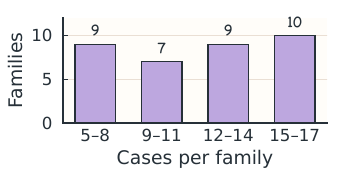}\par
\smallskip
\small\benchsidetitle{(b) Case breadth}\par
\vspace{4pt}
\includegraphics[width=\linewidth]{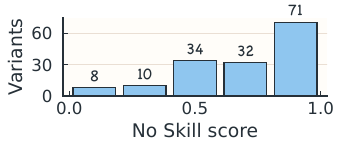}\par
\smallskip
\small\benchsidetitle{(c) Baseline distribution}\par
\vspace{4pt}
\includegraphics[width=\linewidth]{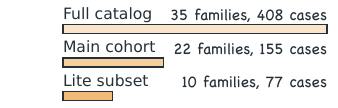}\par
\smallskip
\small\benchsidetitle{(d) Evaluation cohorts}
\end{minipage}
\caption{\textbf{Task-family catalog and evaluation cohorts.}
(a) Domain and family composition, (b) family sizes, and (c) Qwen No-Skill
Avg@3 scores on the \MainVariants{} main-cohort cases. (d) Bar length is proportional to
case counts in the full catalog, main cohort, and lite subset shared by
the additional model pairs.}
\label{fig:benchmark}
\end{figure}

\paragraph{Catalog and evaluation cohorts.}
The \textbf{full catalog} contains \CatalogFamilies{} families and
\CatalogCases{} cases across six domains, with
\CatalogCasesMin{}--\CatalogCasesMax{} cases per family, including the anchor
(Figure~\ref{fig:benchmark}). The \textbf{main cohort} selects families
with at least \MainHeldoutMin{} verified held-out cases and at least two
non-perfect outcomes in catalog-wide No-Skill screening, yielding
\MainFamilies{} families and \MainVariants{} held-out cases
(Table~\ref{tab:family-catalog}); its \textbf{lite subset} covers
\CrossFamilies{} families and \CrossVariants{} cases across all six domains
for comparisons across model pairs. Each pair independently acquires its
own packages.

\subsection{Evaluation and metrics}
\label{sec:benchmark-metrics}

Private checks score each fresh solver's submitted artifacts or state.
$R_{f,v,r}$ is the fraction passed for case $v$ of family $f$ in repeat $r$;
recorded timeouts score zero. All methods share the cases and verifier.
The main study uses \MainRepeats{} independent acquisition runs per family
and condition, each evaluated on all held-out cases; No Skill uses
\MainRepeats{} fresh-solver repeats. Family-macro Avg@3 is
\begin{equation}
 \widehat U=\frac{1}{3|\mathcal F|}
 \sum_{r=1}^{3}\sum_{f\in\mathcal F}
 \frac{1}{|V_f|}\sum_{v\in V_f}R_{f,v,r}.
\end{equation}
Here $\mathcal F$ is the family set and $V_f$ its held-out cases.
Families receive equal weight; additional model pairs use a single repeat.
\textbf{Additional metrics:} domain scores average their families;
transfer gain is $100(\widehat U_{\mathrm{Skill}}-\widehat U_{\mathrm{No\ Skill}})$
percentage points; Qwen main-cohort full-pass rate replaces $R_{f,v,r}$
with $\mathbf{1}[R_{f,v,r}=1]$ in the same average.
\textbf{Uncertainty:} paired 95\% bootstrap intervals resample families
after averaging their cases and repeats; process analyses likewise resample
whole families to retain within-family dependence.

\FloatBarrier
\Needspace{8\baselineskip}
\section{Socratic Skill Learning}
\label{sec:method}

\begin{figure}[!htbp]
\centering
\includegraphics[width=\linewidth]{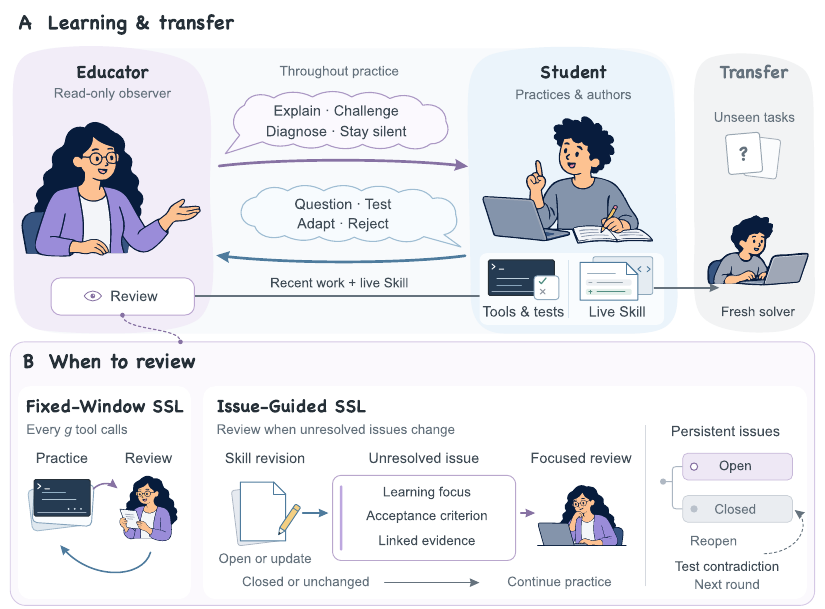}
\caption{\textbf{Socratic Skill Learning.}
\textbf{(A)} The Student tests guidance and authors the Skill; the Educator reviews
recent work and the live package read-only. Only the Skill transfers to fresh solvers.
\textbf{(B)} \fixedwindow reviews every $g$ tool calls; \treedevelop
reviews unresolved issues when they open, change, or reopen, using focused context.}
\label{fig:insight}
\end{figure}

We introduce Socratic Skill Learning (SSL), in which an Educator guides
ongoing investigation while a Student tests the guidance and authors a
transferable Skill. We develop two instantiations: \fixedwindow{}
periodically reviews Student progress, while \treedevelop{} maintains
learning issues to select and focus subsequent reviews. We first describe
their shared learning process (Figure~\ref{fig:insight}A), then present
each protocol (Figure~\ref{fig:insight}B).

\suppressfloats[t]

\subsection{Socratic dialogue and Student practice}
\label{sec:ist-learning}

Both protocols use the target solver's model as the Student, the sole
environment operator and author of $\skillpkg$. Guided by the family
specification, it explores the anchor, tests candidate procedures, and
decides what evidence to gather and what knowledge to retain
(Section~\ref{sec:setup}).

\paragraph{Educator review and Student response.}
The Educator has read-only access to recent Student work and the live Skill.
It can explain, challenge, diagnose, propose checks, or remain silent.
The Student can question, test, adapt, or reject advice, responding through
dialogue, environment experiments, and package revisions. This interaction
examines the assumptions behind a procedure and the evidence needed to
retain it, with the Student deciding how the evidence changes the Skill.

\paragraph{Rounds and transfer.}
A fresh solver tests each round's package on the reset anchor, checking
whether it carries the procedure beyond the author's session. The resulting
trajectory and verifier feedback guide the next round. Student and Educator
sessions persist; evaluation sessions are fresh. Only the final package
transfers to held-out family members.

\subsection{Fixed-Window SSL}
\label{sec:ist-fixed}

Fixed-Window SSL organizes acquisition as repeated periods of Student
practice and Educator review. After every $g$ completed Student tool calls,
the Educator examines recent reasoning and tool results alongside the live
Skill, informed by the family specification, prior lessons, and anchor
feedback. This recurring review can challenge an assumption or suggest a
missing check while an investigation is still open, shaping the evidence
gathered before a procedure is retained.

The Student can clarify or contest a lesson, then resumes exploration and
Skill revision. The resulting
work informs the next review. Silence lets practice continue directly.
The schedule thus fixes when review occurs, while guidance responds to
the Student's progress. The Student determines when to finish the round;
a final review covers any work since the previous review, allowing Student
action before the snapshot (Appendix~\ref{app:protocol-loop}).

\subsection{Issue-Guided SSL}
\label{sec:ist-issues}

Teaching demand changes as a Skill develops: practice can settle a
question, while new evidence can call for further guidance. Issue-Guided
SSL tracks unresolved questions across revisions to select when review
is needed and focus it on relevant evidence.

\paragraph{Persistent learning issues.}
An Educator lesson supplies a learning focus and an acceptance criterion:
what to investigate and what observable evidence would resolve it. A learning
issue retains these together with evidence and package revisions for
subsequent review. Issues form a persistent
\issuetree{}: a concern arising within an investigation becomes a child;
closing it returns attention to the nearest unresolved ancestor. Resolved
branches preserve what has already been addressed.

\paragraph{Revisions and resolution.}
The Student groups file edits into a coherent revision, declaring its
summary, evidence, and any issue addressed. An addressed issue can close
with a changed package when the Student verifies the revision, documents
the procedure's applicability with evidence, or rejects the guidance on
evidential grounds. Verified closure requires a successful post-lesson
check. Investigation can therefore settle a question without another
Educator call.

\paragraph{Selective review and reopening.}
An initial review occurs at the first revision or after a warm-up window
of tool calls. Thereafter, an unresolved issue must open, change, or reopen
before further review. The Educator receives the active issue and its
ancestors, the latest revision, and recent local work, with read access to
the complete package. A lesson returns the Student to investigation;
silence closes the current review episode. Contradictory
fresh-solver anchor feedback can reopen the associated issue in the next
round while preserving resolved siblings.
Appendix~\ref{app:protocol-loop} gives event rules and boundary handling.

\section{Can SSL Develop Transferable Agent Skills?}
\label{sec:experiments}
\label{sec:evaluation}
To assess how teaching supports Skill acquisition, we investigate four
research questions (RQs): \textbf{RQ1} evaluates transfer to held-out family
members; \textbf{RQ2} compares how expert guidance is organized across model
pairs; \textbf{RQ3} traces teaching, Student actions, and package revisions
during acquisition; and \textbf{RQ4} examines whether issue-guided review
preserves transfer while reducing expert supervision. Evaluations follow
the shared protocol in Section~\ref{sec:benchmark}.

\Needspace{7\baselineskip}
\subsection{RQ1: Transfer to held-out family members}
\label{sec:results}

\textit{Do Skills acquired through interleaved teaching transfer to unseen
members of the same task families?}

Table~\ref{tab:domain-transfer} compares overall and domain-level transfer on
the Qwen \MainFamilies{}-family, \MainVariants{}-case main cohort.
Qwen3.7 Plus serves as Student and fresh solver; Qwen3.7 Max supplies
external expertise as Educator or direct package author.
We compare the two SSL schedules with No Skill, direct Expert
Authoring, Post-hoc Teaching, and adapted Trace2Skill and SKILL-KD.
SSL, Expert Authoring, and Post-hoc Teaching use \MainRounds{} acquisition
rounds, with $g=10$ for Fixed-Window SSL. Appendix~\ref{app:implementation}
details model settings, baseline adaptations, and session boundaries;
Appendix~\ref{app:main-family-results} provides per-family results,
case-level gains and regressions, and uncertainty estimates.

The two SSL schedules rank first and second overall and in each of the
six domains. Fixed-Window SSL exceeds adapted Trace2Skill by
\DevelopOverTrace{} percentage points (95\% paired family-bootstrap CI:
$\DevelopOverTraceCI$) and adapted SKILL-KD by
\DevelopOverSkillKD{} points. These gains measure package reuse beyond
the anchor and acquisition session.

The utility improvement extends to complete task solutions. Family-macro
full-pass rate, requiring every private check to pass, rises from
\num[round-mode=places,round-precision=1]{\BaseFullPass}\% without a Skill
to \num[round-mode=places,round-precision=1]{\DevelopFullPass}\% with
Fixed-Window SSL and
\num[round-mode=places,round-precision=1]{\IssueTreeFullPass}\% with
Issue-Guided SSL.

\begin{table}[!htbp]
\centering
\small
\caption{\textbf{Overall and domain transfer on the Qwen main cohort.}
Family-macro Avg@3 utilities across \MainFamilies{} families and \MainVariants{} cases
on a 0--100 scale; superscripts show gain over No Skill (pp).
Domains: Sys. (systems),
Serv. (services), Data (data processing), Sci. (scientific computing),
Fin. (finance), and Office (office automation).
Bold and underlined scores mark the best and second-best learning methods
in each column.}
\label{tab:domain-transfer}
\begingroup
\definecolor{resultGain}{HTML}{666666}
\definecolor{resultStudentRow}{HTML}{EDF5FB}
\definecolor{resultTreeRow}{HTML}{EFF8F5}
\definecolor{resultCompile}{HTML}{F2C887}
\definecolor{resultTeach}{HTML}{BDA7DF}
\definecolor{resultTrace2Skill}{HTML}{EAA6B7}
\definecolor{resultSKILLKD}{HTML}{C7B6A8}
\definecolor{resultDevelop}{HTML}{8FC6EF}
\definecolor{resultTree}{HTML}{84D4C4}
\color{black}
\arrayrulecolor{black}
\fontsize{9}{10.5}\selectfont
\setlength{\tabcolsep}{3pt}
\renewcommand{\arraystretch}{1.17}
\begin{tabularx}{\linewidth}{l*{7}{>{\centering\arraybackslash}X}}
\toprule
Skill method & All & Sys. & Serv. & Data & Sci. & Fin. & Office \\
\midrule
No Skill score & 72.3 & 78.6 & 58.2 & 68.1 & 66.8 & 78.0 & 90.4 \\
\midrule
\compileafter & 72.8\raisebox{.5ex}{\fontsize{7}{8}\selectfont\textcolor{resultGain}{\num{+0.5}}} & 77.9\raisebox{.5ex}{\fontsize{7}{8}\selectfont\textcolor{resultGain}{\num{-0.6}}} & 58.7\raisebox{.5ex}{\fontsize{7}{8}\selectfont\textcolor{resultGain}{\num{+0.6}}} & 71.6\raisebox{.5ex}{\fontsize{7}{8}\selectfont\textcolor{resultGain}{\num{+3.5}}} & 72.4\raisebox{.5ex}{\fontsize{7}{8}\selectfont\textcolor{resultGain}{\num{+5.7}}} & 72.5\raisebox{.5ex}{\fontsize{7}{8}\selectfont\textcolor{resultGain}{\num{-5.5}}} & 91.4\raisebox{.5ex}{\fontsize{7}{8}\selectfont\textcolor{resultGain}{\num{+1.0}}} \\
\teachafter & 74.8\raisebox{.5ex}{\fontsize{7}{8}\selectfont\textcolor{resultGain}{\num{+2.4}}} & 73.6\raisebox{.5ex}{\fontsize{7}{8}\selectfont\textcolor{resultGain}{\num{-4.9}}} & 64.4\raisebox{.5ex}{\fontsize{7}{8}\selectfont\textcolor{resultGain}{\num{+6.3}}} & 73.7\raisebox{.5ex}{\fontsize{7}{8}\selectfont\textcolor{resultGain}{\num{+5.6}}} & 76.4\raisebox{.5ex}{\fontsize{7}{8}\selectfont\textcolor{resultGain}{\num{+9.7}}} & 78.1\raisebox{.5ex}{\fontsize{7}{8}\selectfont\textcolor{resultGain}{\num{+0.0}}} & 91.3\raisebox{.5ex}{\fontsize{7}{8}\selectfont\textcolor{resultGain}{\num{+0.9}}} \\
\tracebaseline & 76.1\raisebox{.5ex}{\fontsize{7}{8}\selectfont\textcolor{resultGain}{\num{+3.8}}} & 85.0\raisebox{.5ex}{\fontsize{7}{8}\selectfont\textcolor{resultGain}{\num{+6.4}}} & 63.4\raisebox{.5ex}{\fontsize{7}{8}\selectfont\textcolor{resultGain}{\num{+5.2}}} & 71.0\raisebox{.5ex}{\fontsize{7}{8}\selectfont\textcolor{resultGain}{\num{+2.9}}} & 73.3\raisebox{.5ex}{\fontsize{7}{8}\selectfont\textcolor{resultGain}{\num{+6.5}}} & 77.9\raisebox{.5ex}{\fontsize{7}{8}\selectfont\textcolor{resultGain}{\num{-0.1}}} & 90.0\raisebox{.5ex}{\fontsize{7}{8}\selectfont\textcolor{resultGain}{\num{-0.4}}} \\
\skillkdbaseline & 75.1\raisebox{.5ex}{\fontsize{7}{8}\selectfont\textcolor{resultGain}{\num{+2.8}}} & 85.5\raisebox{.5ex}{\fontsize{7}{8}\selectfont\textcolor{resultGain}{\num{+6.9}}} & 60.9\raisebox{.5ex}{\fontsize{7}{8}\selectfont\textcolor{resultGain}{\num{+2.7}}} & 73.2\raisebox{.5ex}{\fontsize{7}{8}\selectfont\textcolor{resultGain}{\num{+5.1}}} & 68.2\raisebox{.5ex}{\fontsize{7}{8}\selectfont\textcolor{resultGain}{\num{+1.5}}} & 75.7\raisebox{.5ex}{\fontsize{7}{8}\selectfont\textcolor{resultGain}{\num{-2.4}}} & 90.3\raisebox{.5ex}{\fontsize{7}{8}\selectfont\textcolor{resultGain}{\num{-0.1}}} \\
\rowcolor{resultStudentRow}
\textbf{\fixedwindow} & \underline{80.1}\raisebox{.5ex}{\fontsize{7}{8}\selectfont\textcolor{resultGain}{\num{+7.8}}} & \cellcolor{resultDevelop!34!resultStudentRow}\textbf{88.7}\raisebox{.5ex}{\fontsize{7}{8}\selectfont\textcolor{resultGain}{\num{+10.1}}} & \underline{67.3}\raisebox{.5ex}{\fontsize{7}{8}\selectfont\textcolor{resultGain}{\num{+9.1}}} & \underline{75.1}\raisebox{.5ex}{\fontsize{7}{8}\selectfont\textcolor{resultGain}{\num{+6.9}}} & \underline{77.0}\raisebox{.5ex}{\fontsize{7}{8}\selectfont\textcolor{resultGain}{\num{+10.2}}} & \underline{82.4}\raisebox{.5ex}{\fontsize{7}{8}\selectfont\textcolor{resultGain}{\num{+4.4}}} & \underline{94.1}\raisebox{.5ex}{\fontsize{7}{8}\selectfont\textcolor{resultGain}{\num{+3.7}}} \\
\rowcolor{resultTreeRow}
\textbf{\treedevelop} & \cellcolor{resultTree!34!resultTreeRow}\textbf{80.5}\raisebox{.5ex}{\fontsize{7}{8}\selectfont\textcolor{resultGain}{\num{+8.2}}} & \underline{85.7}\raisebox{.5ex}{\fontsize{7}{8}\selectfont\textcolor{resultGain}{\num{+7.2}}} & \cellcolor{resultTree!34!resultTreeRow}\textbf{71.0}\raisebox{.5ex}{\fontsize{7}{8}\selectfont\textcolor{resultGain}{\num{+12.8}}} & \cellcolor{resultTree!34!resultTreeRow}\textbf{75.9}\raisebox{.5ex}{\fontsize{7}{8}\selectfont\textcolor{resultGain}{\num{+7.8}}} & \cellcolor{resultTree!34!resultTreeRow}\textbf{77.1}\raisebox{.5ex}{\fontsize{7}{8}\selectfont\textcolor{resultGain}{\num{+10.3}}} & \cellcolor{resultTree!34!resultTreeRow}\textbf{83.6}\raisebox{.5ex}{\fontsize{7}{8}\selectfont\textcolor{resultGain}{\num{+5.6}}} & \cellcolor{resultTree!34!resultTreeRow}\textbf{94.6}\raisebox{.5ex}{\fontsize{7}{8}\selectfont\textcolor{resultGain}{\num{+4.2}}} \\
\bottomrule
\end{tabularx}
\endgroup

\end{table}

\Needspace{7\baselineskip}
\subsection{RQ2: Organizing expert guidance during Skill acquisition}
\label{sec:analysis-experiments}

\textit{How should external expertise guide Skill acquisition?}

Table~\ref{tab:ordering}(a) compares the four authoring protocols on the shared
\CrossFamilies{} families across the Qwen, DeepSeek, and Kimi model pairs,
complemented by the Qwen main cohort. Within each pair, the solver and tasks are fixed; protocols
differ in who investigates and writes the package, when guidance is available,
and how the agents interact (Appendices~\ref{app:implementation}
and~\ref{app:main-family-results}).

Guidance during practice consistently produces stronger transfer than
reviewing the Student's completed authoring trajectory. Fixed-Window SSL
and Post-hoc Teaching both preserve the Student as environment operator
and package author, but Fixed-Window SSL outperforms Post-hoc Teaching on all three
model pairs. On the main cohort, this advantage is \DevelopOverTeach{}
percentage points in utility with \ISTPosthocCostReduction{}\% fewer
acquisition tokens.
Interleaving lets the Student test advice while an investigation is still
open and use the resulting evidence in subsequent exploration and writing.
Post-hoc review arrives after those earlier choices have shaped the package.
The results favor feedback that can redirect evidence gathering throughout
practice.

The ordering also distinguishes guided practice from simply retaining a
Student author. On the shared families, Post-hoc Teaching trails Expert
Authoring in every model pair, whereas both interleaved protocols surpass
Expert Authoring. This ordering favors coupling Student authorship with
expert guidance during investigation. RQ3 examines how this interaction
appears in Student actions and package revisions.

\begin{table}[!htbp]
\centering
\caption{\textbf{Cross-model transfer and selective review.}
(a) Utility gain over No Skill (pp) on the shared lite subset
(\CrossFamilies{} families, \CrossVariants{} cases).
(b) Utility and Educator savings: Qwen main cohort
(\MainFamilies{} families, \MainVariants{} cases); DeepSeek and Kimi lite subset.
Savings and efficiency ratios compare Issue-Guided with Fixed-Window SSL.
Qwen uses three repeats; DeepSeek and Kimi use one.
Model pairs are defined in Appendix~\ref{app:model-settings}.}
\label{tab:ordering}
\label{tab:cross-schedule}
\definecolor{orderStudentRow}{HTML}{EDF5FB}
\definecolor{orderTreeRow}{HTML}{EFF8F5}
\begin{minipage}[t]{.477\linewidth}\vspace{0pt}
\centering\small
\textbf{(a) Protocol comparison}\par\smallskip
\setlength{\tabcolsep}{3pt}\renewcommand{\arraystretch}{1.14}
\begin{tabularx}{\linewidth}{@{}>{\raggedright\arraybackslash}Xccc@{}}
\toprule
Protocol & Qwen & DeepSeek & Kimi \\
\midrule
\rowcolor{orderStudentRow}
\fixedwindow & \num{+4.1} & \textbf{\num{+3.1}} & \textbf{\num{+8.2}} \\
\rowcolor{orderTreeRow}
\treedevelop & \textbf{\num{+4.5}} & \num{+2.5} & \num{+7.4} \\
\compileafter & \num{+1.1} & \num{+1.2} & \num{+5.8} \\
\teachafter & \num{-0.7} & \num{+0.5} & \num{+4.1} \\
\bottomrule
\end{tabularx}

\end{minipage}\hfill
\begin{minipage}[t]{.493\linewidth}\vspace{0pt}
\centering\small
\textbf{(b) Selective review}\par\smallskip
\setlength{\tabcolsep}{3pt}\renewcommand{\arraystretch}{1.14}
\begin{tabularx}{\linewidth}{@{}>{\raggedright\arraybackslash}Xccc@{}}
\toprule
Metric & Qwen & DeepSeek & Kimi \\
\midrule
\rowcolor{orderStudentRow}
Fixed-window utility & 0.801 & 0.784 & 0.762 \\
\rowcolor{orderTreeRow}
Issue-guided utility & 0.805 & 0.779 & 0.753 \\
Educator saved (\%) & \TreeEducatorReduction{} & \DeepSeekEducatorReduction{} & \KimiEducatorReduction{} \\
Efficiency ratio & \TreeEducatorEfficiencyRatio{} & \DeepSeekTeachingEfficiencyRatio{} & \KimiTeachingEfficiencyRatio{} \\
\bottomrule
\end{tabularx}

\end{minipage}
\end{table}

\subsection{RQ3: How teaching enters Skill development}
\label{sec:trajectory-findings}
\label{sec:intent-results}
\label{sec:analysis}
\label{sec:case-studies}

\textit{How do teaching, Student actions, and package revisions change as
Skill acquisition progresses?}

\begin{figure}[!htbp]
\centering
\begin{minipage}[t]{0.3091\linewidth}\vspace{0pt}
\centering
\includegraphics[width=\linewidth]{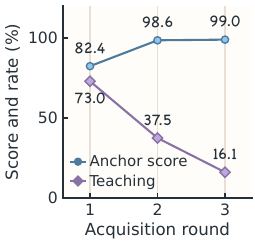}\par
\smallskip
\makebox[\linewidth][l]{\hspace*{.6176\linewidth}\makebox[0pt][c]{\small\textbf{(a) Progress and teaching}}}
\end{minipage}\hfill
\begin{minipage}[t]{0.3091\linewidth}\vspace{0pt}
\centering
\includegraphics[width=\linewidth]{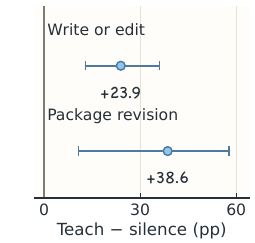}\par
\smallskip
\makebox[\linewidth][l]{\hspace*{.5559\linewidth}\makebox[0pt][c]{\small\textbf{(b) Writing and revision}}}
\end{minipage}\hfill
\begin{minipage}[t]{0.3618\linewidth}\vspace{0pt}
\centering
\includegraphics[width=\linewidth]{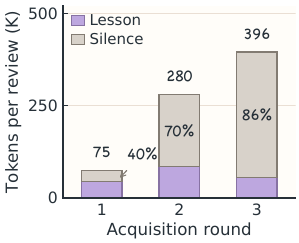}\par
\smallskip
\makebox[\linewidth][l]{\hspace*{.6005\linewidth}\makebox[0pt][c]{\small\textbf{(c) Review cost}}}
\end{minipage}
\caption{\textbf{Teaching demand, Student development, and review cost.}
Qwen Fixed-Window SSL ($g=10$), \ReviewFamilies{} families and \ReviewRuns{} acquisition runs.
Panels show (a) anchor scores and teaching rates at regular reviews,
(b) teaching--silence differences in writing on the next tool call and
Skill changes over ten calls, standardized by round with 95\% family-bootstrap
intervals, and (c) tokens per review before follow-up, with silent shares.}
\label{fig:review-behavior}
\end{figure}

Figure~\ref{fig:review-behavior} traces \ReviewTotal{} reviews across
\ReviewRuns{} Qwen \fixedwindow{} runs with \MainRounds{} rounds each,
linking Educator-reported intents, Student actions, and package snapshots. We compare
behavior following teaching and silence at regular reviews, standardizing
by round. Appendices~\ref{app:intents} and~\ref{app:review-behavior} detail the taxonomy
and estimation.

Teaching demand falls while Skill development continues. As anchor utility
improves, the fraction of regular reviews delivering a lesson falls from
\ReviewFirstRoundRate{}\% in round one to \ReviewLastRoundRate{}\% in round
three, and advice shifts from reasoning toward diagnosis. Under the
three-round protocol, packages still change in \ProgressRevisedAfterPass{}
of \PostPerfectTransitions{} rounds following a perfect anchor score,
including rounds with no new lesson.

Compared with silence, lessons are followed by a
\num[retain-explicit-plus=false]{\ReviewWriteAdjustedPP}-point higher
probability of file writing or editing on the next call and a
\num[retain-explicit-plus=false]{\ReviewPackageAdjustedPP}-point higher
probability of a Skill change over the next ten calls, after adjustment for
round (Figure~\ref{fig:review-behavior}b).

\Needspace{8\baselineskip}
\begingroup
\setlength{\intextsep}{3pt}
\begin{wraptable}{r}{.47\textwidth}
\centering
\tableappearance
\tablecaptionspacing
\caption{\textbf{Testing teaching advice.}
Qwen Fixed-Window SSL, HTML-finance family, repeat 1.
Recorded anchor checks; the erroneous report entry is line \CaseCorrectLine{}.}
\label{tab:case-test}
\small
\setlength{\tabcolsep}{3pt}
\renewcommand{\arraystretch}{1.10}
\begin{tabular}{@{}lrr@{}}
\toprule
Verifier & Flagged & Selected line \\
\midrule
Generic arithmetic & 4 & 140 \\
Financial relations & 1 & \textbf{210} \\
\bottomrule
\end{tabular}

\end{wraptable}

The Qwen HTML-finance case shows how Student checks refine teaching advice.
The Educator proposes deriving reported values from other entries to detect
inconsistency. The Student's initial implementation flags
\CaseInitialFlags{} metrics where only one is wrong. After inspecting this
output, the Student replaces generic arithmetic with financial relations
and reverse derivations, isolating the actual error (Table~\ref{tab:case-test}).
The final Skill retains these relations and distinguishes an untested value
from a contradicted one. On an unseen cash-flow report, a fresh solver reads
the package, checks accounting relations, and passes
\CaseFreshPassed{}/\CaseFreshTotal{} checks (Appendix~\ref{app:chains}).
The lesson initiates an investigation; the Student's execution evidence
determines how the proposal becomes reusable knowledge.
\par\endgroup

Fixed review intervals become increasingly costly as this need for guidance
declines. Average tokens per review grow \ReviewCostGrowth{}-fold from the
first to the last round as context accumulates; reviews returning silence
account for \ReviewLastSilentTokenShare{}\% of final-round review tokens,
before follow-up dialogue. This mismatch motivates selecting reviews around
unresolved work while allowing the Student's investigation to continue.

\FloatBarrier
\Needspace{5\baselineskip}
\subsection{RQ4: Selecting and focusing review around unresolved issues}
\label{sec:efficiency-results}

\textit{Can selecting and focusing review around unresolved issues preserve
transfer while reducing expert supervision?}

Figure~\ref{fig:results-overview} compares fixed reviews every $g=5,10,20$
Student tool calls with \treedevelop{} on the Qwen main cohort, using
\MainRounds{} rounds and \MainRuns{} runs per schedule.
We prioritize Educator tokens to measure demand for external expertise,
and track how Student work changes total acquisition cost.
Table~\ref{tab:cross-schedule}(b) extends the comparison to DeepSeek and Kimi
on the shared \CrossFamilies{} families. Appendix~\ref{app:cost-accounting}
defines the cost components for each model pair.

\begingroup
\setlength{\intextsep}{8.85pt plus 2pt minus 2pt}
\begin{figure}[!htbp]
\centering
\begin{minipage}[t]{0.416364\linewidth}\vspace{0pt}
\hspace*{0.4716\linewidth}\small\textbf{(a) Utility}\par\vspace{11.5pt}
\includegraphics[width=\linewidth]{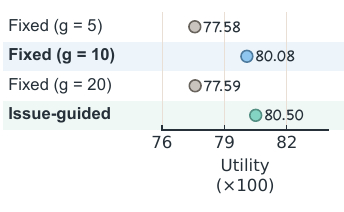}
\end{minipage}%
\begin{minipage}[t]{0.272727\linewidth}\vspace{0pt}
\hspace*{0.0933\linewidth}\small\textbf{(b) Teaching cost}\par\vspace{11.5pt}
\includegraphics[width=\linewidth]{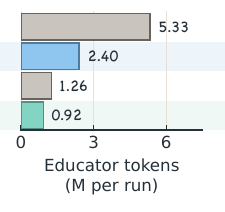}
\end{minipage}%
\begin{minipage}[t]{0.310909\linewidth}\vspace{0pt}
\hspace*{0.1111\linewidth}\small\textbf{(c) Gain per cost}\par\vspace{11.5pt}
\includegraphics[width=\linewidth]{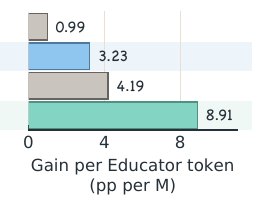}
\end{minipage}
\caption{\textbf{Transfer utility and teaching efficiency across review schedules.}
Qwen main cohort: \MainFamilies{} families, \MainVariants{} cases, \MainRepeats{} repeats.
Panels compare (a) family-macro utility, (b) Educator tokens, and
(c) transfer gain over No Skill per million Educator tokens,
with shared schedule rows.}
\label{fig:results-overview}
\end{figure}
\endgroup

Issue-guided review reaches \TreeUtility{} utility versus \GTenUtility{}
for the standard $g=10$ window, using \TreeEducatorReduction{}\% fewer
Educator tokens and \TreeTotalCostReduction{}\% fewer acquisition tokens.
It achieves \TreeEducatorEfficiencyRatio{} times the transfer gain per
Educator token.

Simply changing the interval does not recover this balance:
$g=5$ increases Educator cost, whereas $g=20$ reduces it, but both yield
lower utility. Compared with $g=10$, Issue-Guided SSL reduces review count
by \ReviewReduction{}\% and tokens per review (including follow-up dialogue)
by \ReviewTokenReduction{}\%.

\Needspace{7\baselineskip}
On DeepSeek and Kimi, Issue-Guided SSL uses \DeepSeekEducatorReduction{}\%
and \KimiEducatorReduction{}\% fewer Educator tokens, respectively, while
retaining utility within one percentage point of Fixed-Window SSL.
Total acquisition tokens also fall for Kimi; additional Student work raises
the total for DeepSeek (Appendix~\ref{app:timing}).

\FloatBarrier
\Needspace{5\baselineskip}
\section{Discussion and Conclusion}
\label{sec:discussion}

On shared \skilltransferbench{} families across three model pairs, Socratic
Skill Learning (\developalong) improves transfer over direct expert authoring
and post-hoc teaching. Process and case analyses show how guidance enters
Skill development: advice identifies a direction to investigate, while
execution evidence informs how the Student tests, adapts, and retains a
procedure.

As teaching demand changes, unresolved questions provide a basis for
allocating expert attention. Issue-Guided SSL tracks these questions across
revisions, retaining transfer utility with \TreeEducatorEfficiencyRatio{}
times the gain per Educator token of Fixed-Window SSL on the main cohort.
The study connects the development of reusable knowledge with decisions
about where further guidance is needed.

\section*{Reproducibility Statement}

Benchmark tasks, acquisition and evaluation code, and released trajectories
are available at \SSLRepositoryLink.
Appendix~\ref{app:assets} documents benchmark construction and evaluation
cohorts; Appendix~\ref{app:implementation} specifies model settings,
acquisition protocols, session boundaries, and reproduction commands.
Appendices~\ref{app:review-behavior} and~\ref{app:timing} detail the process
analyses and token accounting.

\section*{AI Use Statement}

We used generative AI tools to assist with drafting and adapting benchmark
materials, implementing experimental code and analysis scripts, and
analyzing recorded experimental results. We also used these tools for
literature search, manuscript editing, and figure preparation. Benchmark
materials were developed with human experts in coding and agent benchmarking.
Each case underwent programmatic checks and review by an expert other than
its author, following Appendix~\ref{app:assets}. Numerical summaries were
checked against recorded runs and analysis outputs. The authors take
responsibility for the final content.

\setlength{\textfloatsep}{\papertextfloatsep}
\setlength{\floatsep}{\paperfloatsep}
\setlength{\intextsep}{\paperintextsep}
\setlength{\abovecaptionskip}{\papercaptionskip}

\bibliography{references}

\begin{thebibliography}{27}
\providecommand{\natexlab}[1]{#1}
\providecommand{\url}[1]{\texttt{#1}}
\expandafter\ifx\csname urlstyle\endcsname\relax
  \providecommand{\doi}[1]{doi: #1}\else
  \providecommand{\doi}{doi: \begingroup \urlstyle{rm}\Url}\fi

\bibitem[{Anthropic}(2025)]{anthropic2025skills}
{Anthropic}.
\newblock Agent skills overview.
\newblock
  \url{https://platform.claude.com/docs/en/agents-and-tools/agent-skills/overview},
  2025.
\newblock Accessed 2026-08-02.

\bibitem[Chen et~al.(2026)Chen, Wu, Hou, Tang, Hu, et~al.]{agentifoneday2026}
Kaiyuan Chen, Qimin Wu, Taiyu Hou, Tianhao Tang, Xueyu Hu, et~al.
\newblock {AgentIF-OneDay}: A task-level instruction-following benchmark for
  general {AI} agents in daily scenarios.
\newblock \emph{arXiv preprint arXiv:2601.20613}, 2026.

\bibitem[Chi et~al.(2001)Chi, Siler, Jeong, Yamauchi, and
  Hausmann]{chi2001learning}
Michelene T.~H. Chi, Stephanie~A. Siler, Heisawn Jeong, Takashi Yamauchi, and
  Robert~G. Hausmann.
\newblock Learning from human tutoring.
\newblock \emph{Cognitive Science}, 25\penalty0 (4):\penalty0 471--533, 2001.

\bibitem[Guan et~al.(2026)Guan, Wang, Yang, Cao, Liu, Hu, Li, and
  Zhang]{guan2026continualskillbench}
Tianyi Guan, Yiding Wang, Haotong Yang, Siyuan Cao, Shirui Liu, Yi~Hu, Jiaqi
  Li, and Muhan Zhang.
\newblock {ContinualSkillBench}: Can {LLM} agents truly evolve their
  capabilities?
\newblock \emph{arXiv preprint arXiv:2608.03874}, 2026.

\bibitem[Hattie and Timperley(2007)]{hattie2007feedback}
John Hattie and Helen Timperley.
\newblock The power of feedback.
\newblock \emph{Review of Educational Research}, 77\penalty0 (1):\penalty0
  81--112, 2007.

\bibitem[Jiang and Huang(2026)]{agentbrew2026}
Yangqin Jiang and Chao Huang.
\newblock {AgentBrew}: Lifelong knowledge brewing from strong teachers to weak
  {LLM} agents.
\newblock \emph{arXiv preprint arXiv:2607.16851}, 2026.

\bibitem[Li et~al.(2026{\natexlab{a}})Li, Yao, Zhu, Feng, Ye, Wang, He, Zou,
  Zhang, Lei, Huang, Deng, Sun, Zhang, Ye, and Liu]{li2026codetracer}
Han Li, Yifan Yao, Letian Zhu, Rili Feng, Hongyi Ye, Jiaming Wang, Yancheng He,
  Pengyu Zou, Lehan Zhang, Xinping Lei, Haoyang Huang, Ken Deng, Ming Sun,
  Zhaoxiang Zhang, He~Ye, and Jiaheng Liu.
\newblock {CodeTracer}: Towards traceable agent states.
\newblock \emph{arXiv preprint arXiv:2604.11641}, 2026{\natexlab{a}}.

\bibitem[Li et~al.(2026{\natexlab{b}})Li, Liu, Chen, You, Di,
  et~al.]{skillsbench2026}
Xiangyi Li, Yimin Liu, Wenbo Chen, Bingran You, Zonglin Di, et~al.
\newblock {SkillsBench}: Benchmarking how well agent skills work across diverse
  tasks.
\newblock \emph{arXiv preprint arXiv:2602.12670}, 2026{\natexlab{b}}.

\bibitem[Li et~al.(2022)Li, Choi, Chung, Kushman, Schrittwieser,
  et~al.]{codecontests2022}
Yujia Li, David Choi, Junyoung Chung, Nate Kushman, Julian Schrittwieser,
  et~al.
\newblock Competition-level code generation with {AlphaCode}.
\newblock \emph{Science}, 378\penalty0 (6624):\penalty0 1092--1097, 2022.

\bibitem[Li and Hu(2026)]{skillhone2026}
Zhiwei Li and Yong Hu.
\newblock {SkillHone}: A harness for continual agent skill evolution through
  persistent decision history.
\newblock \emph{arXiv preprint arXiv:2606.08671}, 2026.

\bibitem[Liu et~al.(2026{\natexlab{a}})Liu, Ming, Joty, and Zhao]{hasp2026}
Hongjun Liu, Yifei Ming, Shafiq Joty, and Chen Zhao.
\newblock Harnessing {LLM} agents with skill programs.
\newblock \emph{arXiv preprint arXiv:2605.17734}, 2026{\natexlab{a}}.

\bibitem[Liu et~al.(2024{\natexlab{a}})Liu, Huang, Xiao, Sha, Wu, Liu, Wang,
  and Chen]{socraticlm2024}
Jiayu Liu, Zhenya Huang, Tong Xiao, Jing Sha, Jinze Wu, Qi~Liu, Shijin Wang,
  and Enhong Chen.
\newblock {SocraticLM}: Exploring socratic personalized teaching with large
  language models.
\newblock In \emph{Advances in Neural Information Processing Systems},
  volume~37, 2024{\natexlab{a}}.
\newblock
  \url{https://papers.neurips.cc/paper_files/paper/2024/hash/9bae399d1f34b8650351c1bd3692aeae-Abstract-Conference.html}.

\bibitem[Liu et~al.(2024{\natexlab{b}})Liu, Zhu, Liu, Xin, Li, Long, Chen,
  Yang, Xia, Peng, Liu, Zhang, Zhang, Huang, Shen, and
  Xiang]{fullstackbench2024}
Siyao Liu, He~Zhu, Jerry Liu, Shulin Xin, Aoyan Li, Rui Long, Li~Chen, Jack
  Yang, Jinxiang Xia, Z.~Y. Peng, Shukai Liu, Zhaoxiang Zhang, Ge~Zhang, Wenhao
  Huang, Kai Shen, and Liang Xiang.
\newblock {FullStack Bench}: Evaluating {LLM}s as full stack coders.
\newblock \emph{arXiv preprint arXiv:2412.00535}, 2024{\natexlab{b}}.

\bibitem[Liu et~al.(2026{\natexlab{b}})Liu, Su, Xie, Zhang, Zong, Guo, Xie, Ji,
  Yim, Luo, Ren, Chenyu, Li, and Song]{skillrevise2026}
Yuxuan Liu, Zhaochen Su, Lingyun Xie, Yuhao Zhang, Qing Zong, Jiahe Guo,
  Zhongwei Xie, Yiyan Ji, Yauwai Yim, Hongyu Luo, Xiyu Ren, Ruan Chenyu, Haoran
  Li, and Yangqiu Song.
\newblock {SkillRevise}: Improving {LLM}-authored agent skills via
  trace-conditioned skill revision.
\newblock \emph{arXiv preprint arXiv:2606.01139}, 2026{\natexlab{b}}.

\bibitem[Ma et~al.(2026)Ma, Huang, Bao, Zhuang, Shukla, Galley, Zhang, and
  Feuerriegel]{skillgen2026}
Yuchen Ma, Yue Huang, Han Bao, Haomin Zhuang, Swadheen Shukla, Michel Galley,
  Xiangliang Zhang, and Stefan Feuerriegel.
\newblock {SkillGen}: Verified inference-time agent skill synthesis.
\newblock \emph{arXiv preprint arXiv:2605.10999}, 2026.

\bibitem[Ma et~al.(2024)Ma, Zhang, Zhang, Yu, Zhang, Zhang, Luo, Wang, and
  Tang]{spreadsheetbench2024}
Zeyao Ma, Bohan Zhang, Jing Zhang, Jifan Yu, Xiaokang Zhang, Xiaohan Zhang,
  Sijia Luo, Xi~Wang, and Jie Tang.
\newblock {SpreadsheetBench}: Towards challenging real world spreadsheet
  manipulation.
\newblock In \emph{Advances in Neural Information Processing Systems},
  volume~37, pages 94871--94908, 2024.
\newblock \doi{10.52202/079017-3007}.

\bibitem[Madaan et~al.(2023)Madaan, Tandon, Gupta, Hallinan, Gao, Wiegreffe,
  Alon, Dziri, Prabhumoye, Yang, et~al.]{madaan2023selfrefine}
Aman Madaan, Niket Tandon, Prakhar Gupta, Skyler Hallinan, Luyu Gao, Sarah
  Wiegreffe, Uri Alon, Nouha Dziri, Shrimai Prabhumoye, Yiming Yang, et~al.
\newblock Self-refine: Iterative refinement with self-feedback.
\newblock In \emph{Advances in Neural Information Processing Systems}, 2023.

\bibitem[Merrill et~al.(2026)Merrill, Shaw, Carlini, Li, Raj,
  et~al.]{terminalbench2026}
Mike~A. Merrill, Alexander~G. Shaw, Nicholas Carlini, Boxuan Li, Harsh Raj,
  et~al.
\newblock {Terminal-Bench}: Benchmarking agents on hard, realistic tasks in
  command line interfaces.
\newblock In \emph{International Conference on Learning Representations}, 2026.

\bibitem[Ni et~al.(2026)Ni, Liu, Liu, Sun, Zhou, Cheng, Wang, Zhao, Jiang, and
  Jiang]{trace2skill2026}
Jingwei Ni, Yihao Liu, Xinpeng Liu, Yutao Sun, Mengyu Zhou, Pengyu Cheng, Dexin
  Wang, Erchao Zhao, Xiaoxi Jiang, and Guanjun Jiang.
\newblock {Trace2Skill}: Distill trajectory-local lessons into transferable
  agent skills.
\newblock \emph{arXiv preprint arXiv:2603.25158}, 2026.

\bibitem[Quan and Liu(2026)]{econskills2026}
Yinzhu Quan and Zefang Liu.
\newblock {EconSkills}: Studying skill transfer and retrieval for web agents on
  live economic data.
\newblock \emph{arXiv preprint arXiv:2609.19523}, 2026.

\bibitem[Shi et~al.(2026)Shi, Dou, Zhu, Tao, Jin, Kang, Zhou, and
  Weng]{skillkd2026}
Qiming Shi, Yibo Dou, Jiawen Zhu, Yulong Tao, Linbo Jin, Zhaolu Kang, Yunfan
  Zhou, and Di~Weng.
\newblock {SKILL-KD}: Contrastive skill distillation for {LLM} agents.
\newblock \emph{arXiv preprint arXiv:2607.28048}, 2026.

\bibitem[Shinn et~al.(2023)Shinn, Cassano, Gopinath, Narasimhan, and
  Yao]{shinn2023reflexion}
Noah Shinn, Federico Cassano, Ashwin Gopinath, Karthik Narasimhan, and Shunyu
  Yao.
\newblock Reflexion: Language agents with verbal reinforcement learning.
\newblock In \emph{Advances in Neural Information Processing Systems}, 2023.

\bibitem[Wang et~al.(2023)Wang, Xie, Jiang, Mandlekar, Xiao, Zhu, Fan, and
  Anandkumar]{wang2023voyager}
Guanzhi Wang, Yuqi Xie, Yunfan Jiang, Ajay Mandlekar, Chaowei Xiao, Yuke Zhu,
  Linxi Fan, and Anima Anandkumar.
\newblock Voyager: An open-ended embodied agent with large language models.
\newblock \emph{arXiv preprint arXiv:2305.16291}, 2023.

\bibitem[Wang et~al.(2024)Wang, Cui, Zhong, Zhang, Yin, Lin, and
  Shang]{officebench2024}
Zilong Wang, Yuedong Cui, Li~Zhong, Zimin Zhang, Da~Yin, Bill~Yuchen Lin, and
  Jingbo Shang.
\newblock {OfficeBench}: Benchmarking language agents across multiple
  applications for office automation.
\newblock \emph{arXiv preprint arXiv:2407.19056}, 2024.

\bibitem[Zhao et~al.(2024)Zhao, Huang, Xu, Lin, Liu, and Huang]{zhao2024expel}
Andrew Zhao, Daniel Huang, Quentin Xu, Matthieu Lin, Yong-Jin Liu, and Gao
  Huang.
\newblock {ExpeL}: {LLM} agents are experiential learners.
\newblock In \emph{Proceedings of the AAAI Conference on Artificial
  Intelligence}, volume~38, pages 19632--19642, 2024.
\newblock \doi{10.1609/aaai.v38i17.29936}.

\bibitem[Zhao et~al.(2026)Zhao, Li, Li, Zhao, Barr, Sarro, and
  Ye]{zhao2026failureprocess}
Xiangxin Zhao, Han Li, Shuaiting Li, Tianyi Zhao, Earl~T. Barr, Federica Sarro,
  and He~Ye.
\newblock Failure as a process: An anatomy of {CLI} coding agent trajectories.
\newblock \emph{arXiv preprint arXiv:2607.09510}, 2026.

\bibitem[Zhong et~al.(2026)Zhong, Lu, Ning, Wan, Feng, Ao, Ribeiro, Dreyer,
  Ammirati, and Xiong]{skilllearnbench2026}
Shanshan Zhong, Yi~Lu, Jingjie Ning, Yibing Wan, Lihan Feng, Yuyi Ao, Leonardo
  F.~R. Ribeiro, Markus Dreyer, Sean Ammirati, and Chenyan Xiong.
\newblock {SkillLearnBench}: Benchmarking continual learning methods for agent
  skill generation on real-world tasks.
\newblock \emph{arXiv preprint arXiv:2604.20087}, 2026.

\end{thebibliography}
\bibliographystyle{assets/plainnat}

\appendix
\clearpage
\section{Positioning of Prior Skill-Learning Methods}
\label{app:rw-positioning}
Table~\ref{tab:related-positioning} compares where feedback enters Skill
acquisition and how package updates are produced
\citep{trace2skill2026,skillgen2026,skillkd2026,skilllearnbench2026}.
These distinctions motivate
the acquisition-protocol and review-schedule comparisons in the main study.
\begin{table}[!htbp]
\centering\small
\caption{Feedback and package updates in representative Skill-learning
protocols. SSL combines external guidance during ongoing practice with
Student-owned tool use and package writing.}
\label{tab:related-positioning}
\setlength{\tabcolsep}{4pt}
\renewcommand{\arraystretch}{1.15}
\begin{tabularx}{\linewidth}{@{}>{\raggedright\arraybackslash}p{.23\linewidth}>{\raggedright\arraybackslash}X>{\raggedright\arraybackslash}X@{}}
\toprule
Protocol & Feedback enters through & Skill update mechanism \\
\midrule
Trace2Skill & Local trajectory analysis & Extract, merge, and apply lessons \\
SkillGen & Candidate verification & Generate, verify, and refine packages \\
SKILL-KD & Teacher--Student trajectory contrast and Student reruns & Consolidation agent proposes and validates patches \\
SkillLearnBench: Teacher Feedback & Advice after failure, using a reference Skill & Learner revises the Skill before retrying \\
\addlinespace
Fixed-Window SSL & Educator review during practice at fixed tool-call intervals & Student tests advice and writes the live package \\
Issue-Guided SSL & Review on unresolved issue transitions with local evidence & Student revisions and anchor feedback update persistent issue state \\
\bottomrule
\end{tabularx}
\end{table}
\FloatBarrier
\Needspace{10\baselineskip}
\section{Benchmark Assets and Evaluation Cohorts}
\label{app:assets}

\subsection{Construction routes}
Experts in coding and agent benchmarking collaborate with agents to construct
the task families. We first convert source tasks to a common format,
reconciling paths, formats, ordering, precision, and boundary conditions
across the task descriptions, fixtures, and scorers. For example, adapting
the WAL anchor preserves the source database fixtures while revising task
metadata and strengthening checks of the recovered output.

In extraction and adaptation, compatible source cases are grouped into
families, and statements and tests are revised to align their requirements.
In seed expansion, a case from the adapted pool seeds a specification of the
shared invariant and permitted variation; experts and agents then develop
additional cases within this scope. Each task includes a description,
public fixtures, a resettable environment, a reference solution, and a
private verifier.

For example, WAL-recovery cases vary encoding, integrity, and recovery depth
while preserving the reconstruction goal. Acquisition uses a fixed
\texttt{v01} anchor in each family.

\subsection{Catalog and variation axes}
Table~\ref{tab:family-catalog} distinguishes the full catalog, including one
anchor per family, from the main held-out cohort. A dash denotes a family
outside that cohort. Table~\ref{tab:variation-axes} lists the public variation
axes for the evaluated families. The catalog contains \CatalogFamilies{} anchors and
\CatalogNonAnchorCases{} non-anchor cases, with \CatalogCasesMin{}--\CatalogCasesMax{} cases per family; family
members are complete executable tasks.

\Needspace{44\baselineskip}
\begingroup\small
\setlength{\tabcolsep}{5pt}
\setlength{\LTcapwidth}{\linewidth}
\begin{longtable}{llrr}
\caption{Curated task-family catalog and main evaluation cohort, showing case counts for each family within domain groupings and totals across the catalog.}
\label{tab:family-catalog}\\
\toprule
Domain & Family & Catalog cases & Main held-out \\
\midrule
\endfirsthead
\multicolumn{4}{l}{\tablename~\thetable{} (continued)}\\
\toprule
Domain & Family & Catalog cases & Main held-out \\
\midrule
\endhead
\multicolumn{4}{r}{Continued on next page}\\
\endfoot
\bottomrule
\endlastfoot
Systems & Deterministic tarball & 12 & 6 \\
 & ACL inheritance & 13 & 6 \\
 & SQLite truncation & 14 & 6 \\
 & DB WAL recovery & 11 & 7 \\
 & Git leak recovery & 13 & 6 \\
 & Recover obfuscated files & 7 & -- \\
 & Shell deobfuscation & 10 & -- \\
 & Service decommissioning & 6 & -- \\
 & Password recovery & 5 & -- \\
 & Vulnerable secret & 6 & -- \\
\midrule
Services & API scheduler & 15 & 10 \\
 & PyPI server & 16 & 10 \\
 & KV-store gRPC & 16 & 6 \\
 & Flask & 9 & 6 \\
 & WASM pipeline & 11 & -- \\
\midrule
Data Processing & Filter JS from HTML & 16 & 10 \\
 & SPARQL professors & 15 & 10 \\
 & Organization JSON & 13 & 8 \\
 & JQ processing & 12 & 7 \\
 & Multi-source merger & 14 & -- \\
 & Cocktail recipes & 8 & -- \\
\midrule
Scientific Computing & Cross-entropy method & 13 & 6 \\
 & Audio STFT peaks & 11 & 10 \\
 & ODE RK4 & 15 & 5 \\
 & Camera layout & 16 & -- \\
\midrule
Finance & Bank transaction filter & 16 & 6 \\
 & Financial documents & 13 & 6 \\
 & HTML finance verification & 16 & 6 \\
 & Financial analysis & 7 & 6 \\
 & Stock analysis & 11 & -- \\
\midrule
Office Automation & Office workflow & 8 & 6 \\
 & XLSX structure & 10 & 6 \\
 & XLSX formula & 17 & -- \\
 & Excel automation & 7 & -- \\
 & PPTX editing & 6 & -- \\
\midrule
\multicolumn{2}{l}{Total: \CatalogFamilies{} catalog families; \MainFamilies{} main families} & \CatalogCases{} & \MainVariants{} \\
\end{longtable}
\endgroup

\subsection{Validation and acceptance}
Acceptance requires a stable family invariant with declared variation,
unambiguous requirements with deterministic tests, tasks that go beyond
mechanical edits, and at least three separately attributable checks.
Programmatic validation verifies Docker builds, executes tests, and
confirms that reference solutions pass them. An expert other than the
author reviews each case before acceptance. Reference solutions and
private tests remain outside agent workspaces.

Curation yields \CatalogFamilies{} families and \CatalogCases{} cases from
\CatalogCandidateFamilies{} candidate families and \CatalogCandidateCases{}
cases. Reference solutions pass all \CatalogReferenceChecks{} checks
(\CatalogChecksMin{}--\CatalogChecksMax{} per case).

\subsection{Evaluation cohort}
The main cohort selects families with at least \MainHeldoutMin{} verified
held-out cases and at least two non-perfect outcomes in catalog-wide
No-Skill screening. It contains \MainFamilies{} families and
\MainVariants{} held-out cases, with \MainHeldoutMin{}--\MainHeldoutMax{}
cases per family, retaining ceiling cases for coverage and regression
measurement.
Appendix~\ref{app:repeat-headroom} reports repeat-level scores and case-level
transfer directions.

\Needspace{31\baselineskip}
\begingroup\small
\setlength{\LTcapwidth}{\linewidth}
\begin{longtable}{@{}>{\raggedright\arraybackslash}p{.25\linewidth}>{\raggedright\arraybackslash}p{\dimexpr.75\linewidth-2\tabcolsep\relax}@{}}
\caption{Public variation axes for each family in the main evaluation cohort, summarizing the changes across cases declared in the family specifications.}
\label{tab:variation-axes}\\
\toprule
Family & What changes across cases \\
\midrule
\endfirsthead
\multicolumn{2}{l}{\tablename~\thetable{} (continued)}\\
\toprule
Family & What changes across cases \\
\midrule
\endhead
\midrule
\multicolumn{2}{r}{Continued on next page}\\
\endfoot
\bottomrule
\endlastfoot
Filter JS from HTML & Executable constructs, preservation requirements, malformed syntax, encodings. \\
API scheduler & Task volume, emergencies, priority rules, dependencies, API capabilities. \\
SPARQL professors & Rank semantics, affiliation paths, graph noise and duplicate bindings. \\
Audio STFT peaks & Signal recipe, STFT settings, peak selection and shell output contract. \\
PyPI server & Package layout, public API, index configuration, publication and installation. \\
Organization JSON & Entity relationships, field normalization, aggregation distributions. \\
DB WAL recovery & WAL encoding, base/frame integrity, recovery depth, schema and update topology. \\
JQ processing & Selection and ordering, reshaping, normalization, collection-level aggregation. \\
Cross-entropy method & Environment dynamics, elite-update regime, shared-prefix plan evaluation. \\
Deterministic tarball & Source contents and exclusions, metadata normalization, archive/compressor format. \\
Flask & Stateful resources, coupled endpoints, validation and failure behavior. \\
KV-store gRPC & RPC schema, persistent state semantics, service lifecycle and request patterns. \\
HTML finance verification & Report structure, accounting relations, error type, output precision. \\
ODE RK4 & Differential dynamics, evaluation schedule, step and efficiency constraints. \\
Bank transaction filter & Noisy identity signals, membership propagation, confusable distractors. \\
Git leak recovery & Object/ref location, repository topology, recovery and cleanup scope. \\
Financial documents & Content-based classification, OCR/modalities, amount expression, batch composition. \\
SQLite truncation & Corruption mechanism, value encoding, record identification, recovery scale. \\
ACL inheritance & User/group policy topology, effective permissions, descendant inheritance. \\
Financial analysis & Source formats, cross-statement joins, metric definition, locale and schema. \\
Office workflow & Source/destination artifacts, temporal routing, cross-output consistency. \\
XLSX structure & Transformation, workbook topology, alignment, output geometry, preservation. \\
\end{longtable}
\endgroup

\FloatBarrier
\Needspace{8\baselineskip}
\section{Protocols, Implementation, and Reproduction}
\label{app:implementation}

\subsection{Runtime, sessions, and package boundaries}
\label{app:model-settings}
\paragraph{Runtime and models.}
We abbreviate the three model pairs as Qwen, DeepSeek, and Kimi. In each
pair, the Student model also serves as the fresh solver, while the Educator
model supplies external expertise or writes the package under Expert
Authoring.
The harness uses Pi coding agent 0.81.1\footnote{\url{https://pi.dev}},
TypeScript 5.9.3, and Docker task environments. Main Qwen runs use
\texttt{qwen3.7-plus} as Student and \texttt{qwen3.7-max} as Educator,
with a 1,000,000-token
context window, a 65,536-token maximum completion, and thinking enabled.
Sampling uses the provider defaults.
DeepSeek uses \texttt{deepseek-v4-flash-0731} as Student and
\texttt{deepseek-v4-pro-0813} as Educator,
with the same context window, 393,216 maximum completion tokens,
thinking enabled, and reasoning effort \texttt{max}.
Kimi uses \texttt{kimi-k2.5} for the Student and fresh solvers
(262,144-token context, 16,384 maximum completion, thinking enabled),
and \texttt{kimi-k3} for the external author or Educator
(1,048,576-token context, 131,072 maximum completion, reasoning effort
\texttt{max}).

\paragraph{Session isolation and access.}
Within an original authoring-protocol run, the package author (the external
model under Expert Authoring; the Student otherwise) and, where applicable,
the Educator maintain separate continuous sessions across acquisition rounds,
with automatic context compaction. Sessions never continue across families,
replicates, or protocol cells. Every anchor and held-out evaluation uses a
fresh target-model solver. Adapted SKILL-KD instead uses fresh executions
with persistent rule-edit history (Appendix~\ref{app:skill-kd}).

The author operates in a Docker workspace containing the public task, anchor
inputs, and family specification, and is the only agent that writes the live
Skill. Educator tools list, read, and search the full package but cannot execute
environment actions or write files. Solvers load only the frozen target package
through the native Skill catalog, with ambient Skills disabled and a clean case
workspace. The package contains \texttt{SKILL.md} and optional
\texttt{scripts/}, \texttt{references/}, and \texttt{assets/} directories.
Held-out materials remain inaccessible during acquisition, and private checks
are applied only after the solver submits its solution. Snapshots retain every
package file; resource
reads and executable invocations are logged.

\subsection{Acquisition loop and review events}
\label{app:protocol-loop}
\paragraph{Review context and interaction.}
\fixedwindow{} reviews after every $g$ completed Student tool calls, with
$g=10$ in the main condition and $g=5,20$ in the schedule comparison.
\fixedwindow{} provides the Educator with the family definition, recent
reasoning and tool calls, prior lessons, feedback from anchor tests by fresh
solvers, and the complete live package. An opening lesson may explain, propose
a test, diagnose, or
challenge an assumption; the Student may clarify, adapt, or contest it,
followed by at most one Educator reply. Subsequent task actions and package
edits remain Student decisions.

\paragraph{Round completion.}
The author ends its trajectory when it has finished the current exploration
and package revision. The harness completes the scheduled terminal review,
if any, and the Student's resulting work, then validates and snapshots the
package for a fresh anchor test. Each round permits at most one terminal
review; resulting Student work is retained without another terminal review.
The tool-call window controls review frequency, while the author determines
trajectory length. All \MainRounds{} rounds run, with anchor feedback entering
the next round.

\begin{figure}[!htbp]
\small
The loop below preserves Student package ownership under both schedules.
\par\smallskip
\begin{verbatim}
create Student; create Educator; initialize package K
for round in 1..3:
  give Student public spec, clean anchor, previous feedback
  while the current authoring trajectory is active:
    let Student reason, invoke tools, and edit K
    Fixed-Window SSL: review after g completed tool calls
      or at an unreviewed end-of-round tail
    Issue-Guided SSL: on bootstrap, revision commit, or contradiction:
      resolve addressed active issue if closure conditions hold
      otherwise update, open, or reopen its persistent issue
      request review only for an unresolved transition
    if review requested:
      Educator observes local evidence and complete package
      if lesson: Student responds
        fixed-window permits one Educator follow-up
      else: retain Issue-Guided SSL issue with no-lesson closure reason
      Student resumes its own work
  freeze K; evaluate with fresh target model on clean anchor
  retain trajectory and verifier feedback for the next round
evaluate final K with a fresh target model for each held-out case
\end{verbatim}
\end{figure}

\paragraph{Revision declarations and issue state.}
During practice, the Student calls \texttt{commit\_skill\_revision} after a
coherent knowledge change to the complete package. The declaration includes
a summary and evidence, plus an issue ID and disposition when it addresses
an existing lesson. Ordinary file edits accumulate until the Student commits
such a revision. The Educator supplies a learning focus and an observable
acceptance criterion when delivering a lesson.

Each persistent issue stores a focus, acceptance criterion, linked revisions,
evidence, and disposition. A child records a concern within ongoing work;
closure returns attention to the nearest unresolved ancestor. In Issue-Guided
SSL, the first-round bootstrap occurs at the first revision or after ten tool
calls, whichever comes first. At the round boundary, an uncommitted package change is processed as
an implicit revision; the runtime also handles any pending bootstrap or anchor
contradiction before the final snapshot. The default Educator view includes up to three active-path nodes, two
recently closed issues, the latest revision, and recent local tool calls;
read access to the complete package remains available. After bootstrap, another
unresolved transition precedes further teaching, so triggers, context, and
follow-up policy change together.

Student-declared closure requires an addressed active issue and a changed
package. A verified disposition additionally requires a successful post-lesson
verification event; bounded and rejected dispositions record stated evidence.
A no-lesson closure records Educator silence rather than verified correctness.
Contradictory feedback from an anchor test by a fresh solver targets the issue
associated with that package and may reopen it while preserving resolved siblings.

\subsection{Acquisition comparators}
Expert Authoring assigns exploration and package writing to the external model,
including feedback from multi-round anchor tests by fresh solvers. Post-hoc
Teaching keeps the Student as author and offers one review after the authoring
trajectory in each round. A lesson is followed by Student checks and revisions;
this protocol has no immediate Educator follow-up. Table~\ref{tab:protocol-roles}
summarizes the interaction structure. No Skill
skips acquisition and solves each case in a fresh session. Our adaptation of the
released Trace2Skill retains parallel patch proposal and consolidation over
independent anchor trajectories generated from a generic seed package, replacing the
spreadsheet-specific front end with a task-family adapter. The main authoring
comparison contains \MainAuthoringRuns{} acquisition runs; Issue-Guided SSL adds \MainRuns{}.

\paragraph{Cost accounting.}
\label{app:cost-accounting}
Acquisition tokens include cached input and exclude held-out solving.
Qwen totals cover the author, Educator, and fresh-solver anchor evaluations;
DeepSeek and Kimi totals cover the persistent author and Educator sessions.
Teaching efficiency divides transfer gain by Educator tokens.

\begin{table}[!htbp]
\centering\small
\caption{Authoring and teaching protocols. Each uses \MainRounds{} acquisition
rounds with fresh-solver anchor feedback between rounds.}
\label{tab:protocol-roles}
\setlength{\tabcolsep}{4pt}
\renewcommand{\arraystretch}{1.12}
\begin{tabularx}{\linewidth}{@{}>{\raggedright\arraybackslash}p{.20\linewidth}>{\raggedright\arraybackslash}p{.16\linewidth}>{\raggedright\arraybackslash}X>{\raggedright\arraybackslash}X@{}}
\toprule
Protocol & Operator / author & Review opportunity & Continuation \\
\midrule
Expert Authoring & External model & No separate Educator & Author revises from anchor feedback \\
Post-hoc Teaching & Student & Once after the authoring trajectory in each round & Student checks and revises \\
Fixed-Window SSL & Student & Every $g$ tool calls and an unreviewed tail & Student responds; one optional Educator follow-up \\
Issue-Guided SSL & Student & Unresolved issue transitions & Student investigates and commits revisions before further review \\
\bottomrule
\end{tabularx}
\end{table}

\subsubsection{Adapted SKILL-KD}
\label{app:skill-kd}
\paragraph{Adaptation.}
We use the released \skillkdbaseline loop~\citep{skillkd2026}\footnote{\url{https://github.com/KhanCold/SKILL-KD}}
at upstream revision
\texttt{f76b51c3e855}, with adapter revision \texttt{54954317354d}; full hashes are
retained in the compact evidence. Each of \MainRuns{} independent runs starts
from a generic rule seed and makes three visits to its public anchor. A
successful Student attempt leaves the rules unchanged. Otherwise, Teacher and
Student trajectories inform a critic's proposed rule edits, accepted only after
a fully successful fresh Student rerun. Each visit permits at most three
adaptive rounds; failed candidates roll back to the visit-start rules.
Qwen3.7 Plus is the Student and held-out solver; Max is Teacher and critic.
Every execution uses a fresh session. Only accepted rules, not private edit
history, enter the frozen Skill.

\paragraph{Separate cost scope.}
The physical-usage audit records \SkillKDRecordedAcquisitionM{}M acquisition
tokens across \MainRuns{} runs (\SkillKDRecordedTokensPerRunM{}M per run). The
acquisition-cost columns show a dash for this baseline, since the audit
follows a different scope.

\Needspace{12\baselineskip}
\subsection{Reproduction}
The released repository contains tasks in \path{benchmark/}, the
execution harness in \path{src/}, and experiment records in \path{experiments/}.
The file \path{configs/scopes/main.yaml} selects the
\MainFamilies{}-family, \MainVariants{}-case main cohort, and
\path{configs/scopes/lite.yaml} selects the shared 10-family subset.
Model pairs are specified in \path{configs/models/}.
Method identifiers match the paper names: \texttt{expert-authoring},
\texttt{post-hoc-teaching}, \texttt{fixed-window-ssl}, and
\texttt{issue-guided-ssl}.

With Node.js 22.19 or later and Docker, run the following commands from the
repository root. Configure \texttt{SSL\_URL} and \texttt{SSL\_KEY} in
\path{.env} before acquisition or evaluation:

\begingroup\small
\noindent\begin{minipage}{\linewidth}
\begin{verbatim}
npm ci
cp .env.example .env
npm run ssl -- run --scope main --model qwen \
  --methods fixed-window-ssl,issue-guided-ssl
npm run ssl -- evaluate --scope main --model qwen \
  --methods fixed-window-ssl,issue-guided-ssl \
  --frozen experiments/main/qwen
\end{verbatim}
\end{minipage}
\endgroup
\par\smallskip
Add \texttt{--dry-run} to inspect a run plan, or use \texttt{acquire} and
\texttt{evaluate} as separate stages. \path{docs/running.md} documents the
\texttt{trace2skill} and \texttt{skill-kd} baseline commands and setup.
\path{experiments/README.md} maps cohorts to teaching sessions, Skill packages,
and evaluation records. Scores and frozen Skills are directly available;
\texttt{python3 scripts/unpack-experiments.py --group main-qwen} restores
the remaining Qwen session logs. Acquisition uses only the
anchor and public family specification; held-out evaluation uses fresh
solver sessions with the resulting package.

\FloatBarrier
\Needspace{10\baselineskip}
\begingroup
\setlength{\intextsep}{7pt plus 1pt minus 1pt}
\setlength{\textfloatsep}{10pt plus 1pt minus 1pt}
\setlength{\floatsep}{8pt plus 1pt minus 1pt}
\section{Supplementary Transfer Results}
\label{app:main-family-results}

\subsection{Overall and per-family comparisons}
Table~\ref{tab:main-overall-detail} gives family-macro Avg@3 utility and
full-pass rate, paired 95\% bootstrap confidence intervals for utility gains,
and mean acquisition cost on the Qwen \MainFamilies{}-family,
\MainVariants{}-case main cohort. A cost
dash denotes no acquisition for No Skill and unavailable common-scope cost for
the adapted baselines.
Table~\ref{tab:family-all} provides the full per-family values underlying
Table~\ref{tab:domain-transfer}.

\begin{table}[!htbp]
\caption{Utility, full-pass rate, and acquisition cost on the Qwen
\MainFamilies{}-family, \MainVariants{}-case main cohort (Avg@3, family-macro).
Gain and 95\% CI refer to utility differences from No Skill.}
\label{tab:main-overall-detail}
\label{tab:main-result}
\centering\footnotesize
\setlength{\tabcolsep}{3pt}
\renewcommand{\arraystretch}{1.10}
\begin{tabular*}{\linewidth}{@{\extracolsep{\fill}}l S[table-format=1.3] S[table-format=+1.2] c S[table-format=2.2] S[table-format=1.2]@{}}
\toprule
Condition & {Utility $\uparrow$} & {Gain (pp)} & {95\% CI} & {\shortstack{Full-pass\\(\%) $\uparrow$}} & {\shortstack{Tokens/run\\(M)}} \\
\midrule
No Skill & \BaseUtility & {--} & -- & \BaseFullPass & {--} \\
\compileafter & \CompileUtility & +\CompileGain & $\CompileCI$ & \CompileFullPass & \CompileTokens \\
\teachafter & \TeachUtility & +\TeachGain & $\TeachCI$ & \TeachFullPass & \TeachTokens \\
Adapted \tracebaseline & \TraceUtility & +\TraceGain & $\TraceCI$ & \TraceFullPass & {--} \\
Adapted \skillkdbaseline & \SkillKDUtility & +\SkillKDGain & $\SkillKDCI$ & \SkillKDFullPass & {--} \\
\fixedwindow{} ($g=10$) & \GTenUtility & +\DevelopGain & $\DevelopCI$ & \DevelopFullPass & \GTenTokens \\
\midrule
\treedevelop & \bfseries \TreeUtility & \bfseries +\IssueTreeGain & $\IssueTreeCI$ & \bfseries \IssueTreeFullPass & \TreeTokens \\
\bottomrule
\end{tabular*}
\end{table}

\begin{table}[!htbp]
\caption{Per-family Avg@3 scores on the Qwen main cohort
(\MainFamilies{} families, \MainVariants{} cases), including adapted Trace2Skill and
SKILL-KD. Only the last row is a macro-average across task families.}
\label{tab:family-all}
\centering
\footnotesize
\setlength{\tabcolsep}{3pt}
\let\resultshortstack\shortstack
\renewcommand{\shortstack}[2][c]{%
  \ifx#1l{\let\\\space #2}\else\resultshortstack[#1]{#2}\fi}
\begin{tabular}{lrrrrrrr}
\toprule
Family & \shortstack{No\\Skill} & \shortstack{Expert\\Authoring} & \shortstack{Post-hoc\\Teaching} & Trace2Skill & SKILL-KD & \shortstack{Fixed-\\Window\\SSL} & \shortstack{Issue-\\Guided\\SSL} \\
\midrule
\ComparisonFamilyRows
\midrule
\ComparisonFamilyMacro
\bottomrule
\end{tabular}
\end{table}

Full-pass gains over No Skill are \DevelopFullPassGain{} percentage points
for \fixedwindow{} (95\% paired family-bootstrap CI:
$\DevelopFullPassCI$) and \IssueTreeFullPassGain{} points for
\treedevelop{} ($\IssueTreeFullPassCI$), using the same family resampling
as the utility comparisons.

Fixed-Window SSL's paired gains are \DevelopOverCompile{} points over Expert Authoring,
\DevelopOverTeach{} over Post-hoc Teaching, \DevelopOverTrace{} over Trace2Skill,
and \DevelopOverSkillKD{} over adapted SKILL-KD.
It improves on No Skill in \DevelopPositiveFamilies{} of the \MainFamilies{}
families and uses \ISTPosthocCostReduction{}\% fewer acquisition tokens than
Post-hoc Teaching.
The 95\% bootstrap confidence interval over task families for Fixed-Window SSL--Trace2Skill is
$\DevelopOverTraceCI$ percentage points. The intervals for Post-hoc Teaching--Expert Authoring,
Fixed-Window SSL--Post-hoc Teaching, and Fixed-Window SSL--Expert Authoring are respectively
$\TeachOverCompileCI$, $\DevelopOverTeachCI$, and $\DevelopOverCompileCI$ points.
The Fixed-Window SSL--adapted SKILL-KD interval is
$\DevelopOverSkillKDCI$ points. All intervals resample the same paired family
differences.

\FloatBarrier
\Needspace{8\baselineskip}
\subsection{Repeat variation, non-ceiling cases, and sensitivity}
\label{app:repeat-headroom}
Table~\ref{tab:variant-directions} compares each case's Avg@3 score with
its No-Skill score. Counts are independent of family or domain weighting.
The ceiling subset contains the \CeilingCases{} cases with No-Skill score 1
across all repeats; the other \HeadroomCases{} cases are non-ceiling. Fixed-Window SSL improves
\DevelopHeadroomPercent\% of non-ceiling cases, and Issue-Guided SSL improves
\IssueTreeHeadroomPercent\%.

\begin{table}[!htbp]
\caption{Case-level Avg@3 transfer directions relative to No Skill on the
Qwen main cohort (\MainFamilies{} families, \MainVariants{} cases), with
non-ceiling and ceiling subsets for Fixed-Window SSL and Issue-Guided SSL.}
\label{tab:variant-directions}
\centering\footnotesize
\begin{tabular}{lrrr}
\toprule
Method & Improved & Unchanged & Regressed \\
\midrule
\multicolumn{4}{l}{All cases ($n=\MainVariants{}$)} \\
\CohortDirectionRows
\SkillKDDirectionRow
\midrule
\multicolumn{4}{l}{Non-ceiling ($n=\HeadroomCases{}$)} \\
Fixed-Window SSL & \DevelopHeadroomRow \\
Issue-Guided SSL & \IssueTreeHeadroomRow \\
\midrule
\multicolumn{4}{l}{Ceiling ($n=\CeilingCases{}$)} \\
Fixed-Window SSL & \DevelopCeilingRow \\
Issue-Guided SSL & \IssueTreeCeilingRow \\
\bottomrule
\end{tabular}
\end{table}
\begin{table}[!htbp]
\caption{Repeat variation on the Qwen main cohort (\MainFamilies{} families,
\MainVariants{} cases): family-macro utility for each repeat, its standard
deviation across repeats, and mean within-family variability.}
\centering\footnotesize
\begin{tabular}{lrrrrr}
\toprule
Condition & Repeat 1 & Repeat 2 & Repeat 3 & SD & Mean within-family SD \\
\midrule
\CohortRepeatRows
\SkillKDRepeatRow
\bottomrule
\end{tabular}
\end{table}

Every Fixed-Window SSL and Issue-Guided SSL repeat exceeds its corresponding No-Skill repeat.
SD is the sample standard deviation across the \MainRepeats{} repeat-level scores. The final column averages the within-family sample SD
across \MainFamilies{} families.

\FloatBarrier
\Needspace{12\baselineskip}
\subsection{Additional model pairs}
\label{app:cross-model}
DeepSeek-V4-Flash-0731 / DeepSeek-V4-Pro-0813 and Kimi K2.5 / Kimi K3
independently acquire packages on a
\CrossFamilies{}-family, \CrossVariants{}-case lite subset of the main
cohort: \CrossCohortSizes. Each condition has \CrossRepeats{} repeat, \CrossIterations{}
acquisition rounds, and \CrossVariants{} held-out evaluations in fresh
sessions with equal weighting across task families. Each model-pair
experiment comprises \CrossAcquisitionRuns{} acquisition runs,
\CrossAnchorEvaluations{} anchor evaluations by fresh solvers,
\CrossSkillEvaluations{} Skill-based held-out evaluations, and
\CrossNoSkillEvaluations{} No-Skill evaluations, and compares four
acquisition protocols and No Skill.

On all three model pairs, the two interleaved protocols take the top two
positions and Expert Authoring ranks third (Table~\ref{tab:ordering}(a)).
Fixed-Window SSL exceeds Post-hoc Teaching by \SubsetDevelopOverTeach{},
\DeepSeekDevelopOverTeach{}, and \KimiDevelopOverTeach{} points on the shared
families for Qwen, DeepSeek, and Kimi, respectively.
Fixed-Window SSL has the highest mean utility in both additional pairs, gaining
\DeepSeekISTGain{} and \KimiISTGain{} points over the respective No-Skill
solvers; Table~\ref{tab:cross-model-utility} gives the complete comparison.
Table~\ref{tab:cross-model-ci} reports paired family bootstrap intervals for
the headline contrasts, and Table~\ref{tab:cross-family-all} lists the
per-family scores.

\begin{table}[!htbp]
\caption{Utility and gain over No Skill for DeepSeek and Kimi on the lite
subset (\CrossFamilies{} families, \CrossVariants{} cases; single repeat).}
\label{tab:cross-model-utility}
\label{tab:cross-model}
\centering\footnotesize
\setlength{\tabcolsep}{5pt}
\renewcommand{\arraystretch}{1.0}
\begin{tabular*}{\linewidth}{@{\extracolsep{\fill}}l rr rr@{}}
\toprule
& \multicolumn{2}{c}{DeepSeek} & \multicolumn{2}{c}{Kimi} \\
\cmidrule(lr){2-3}\cmidrule(l){4-5}
Condition & Utility & \shortstack{Gain\\(pp)} & Utility & \shortstack{Gain\\(pp)} \\
\midrule
No Skill & .754 & -- & .679 & -- \\
\compileafter & .766 & +1.21 & .738 & +5.85 \\
\teachafter & .759 & +0.54 & .721 & +4.15 \\
\fixedwindow & \textbf{.784} & \textbf{+3.08} & \textbf{.762} & \textbf{+8.24} \\
\treedevelop & .779 & +2.50 & .753 & +7.36 \\
\bottomrule
\end{tabular*}

\end{table}

\begin{table}[!htbp]
\caption{Headline contrasts on the shared lite subset (\CrossFamilies{} families,
\CrossVariants{} cases): 95\% paired family-bootstrap intervals for
Table~\ref{tab:ordering}(a). Qwen averages \MainRepeats{} repeats;
DeepSeek and Kimi use one.}
\label{tab:cross-model-ci}
\centering\footnotesize
\setlength{\tabcolsep}{2pt}
\begin{tabular*}{\linewidth}{@{\extracolsep{\fill}}l rrr@{}}
\toprule
Contrast & Qwen (shared) & DeepSeek & Kimi \\
\midrule
\fixedwindow{} $-$ No Skill & [+1.02, +8.03] & [$-$4.63, +13.03] & [+2.48, +15.62] \\
\treedevelop{} $-$ \fixedwindow & [$-$0.84, +1.44] & [$-$5.34, +3.14] & [$-$4.72, +2.89] \\
\fixedwindow{} $-$ \compileafter & [$-$0.83, +7.14] & [$-$1.93, +7.15] & [$-$2.33, +7.98] \\
\compileafter{} $-$ \teachafter & [$-$1.28, +6.05] & [$-$0.89, +2.23] & [$-$1.13, +4.76] \\
\bottomrule
\end{tabular*}

\end{table}

\begin{table}[!htbp]
\caption{Per-family scores for DeepSeek and Kimi on the lite subset
(\CrossFamilies{} families, \CrossVariants{} cases; single repeat).
Qwen scores for these families appear in Table~\ref{tab:family-all}.}
\label{tab:cross-family-all}
\centering\footnotesize
\renewcommand{\arraystretch}{.96}
\setlength{\tabcolsep}{3pt}
\begin{tabular*}{\linewidth}{@{\extracolsep{\fill}}l rrrrr@{}}
\toprule
Family & \shortstack{No\\Skill} & \shortstack{Expert\\Authoring} & \shortstack{Post-hoc\\Teaching} & \shortstack{Fixed-Window\\SSL} & \shortstack{Issue-Guided\\SSL} \\
\midrule
\multicolumn{6}{l}{DeepSeek} \\
Audio STFT peaks & .668 & .657 & .677 & .637 & .570 \\
Filter JS from HTML & .315 & .485 & .452 & .712 & .529 \\
Git leak recovery & .963 & .891 & .891 & .912 & .912 \\
HTML finance verification & .833 & 1.000 & 1.000 & 1.000 & 1.000 \\
JQ processing & .909 & .891 & .846 & .817 & .835 \\
KV-store gRPC & .656 & .635 & .600 & .668 & .689 \\
PyPI server & .558 & .617 & .592 & .642 & .642 \\
SPARQL professors & .718 & .652 & .672 & .652 & .722 \\
SQLite truncation & .972 & .898 & .931 & .875 & .944 \\
XLSX structure & .944 & .933 & .931 & .931 & .944 \\
\midrule
\multicolumn{6}{l}{Kimi} \\
Audio STFT peaks & .450 & .635 & .585 & .660 & .660 \\
Filter JS from HTML & .248 & .366 & .314 & .593 & .456 \\
Git leak recovery & .867 & .891 & .915 & .891 & .888 \\
HTML finance verification & .833 & 1.000 & .917 & .889 & 1.000 \\
JQ processing & .826 & .817 & .846 & .891 & .864 \\
KV-store gRPC & .607 & .635 & .635 & .635 & .635 \\
PyPI server & .542 & .558 & .600 & .583 & .633 \\
SPARQL professors & .592 & .647 & .652 & .677 & .622 \\
SQLite truncation & .907 & .913 & .812 & .866 & .838 \\
XLSX structure & .921 & .917 & .933 & .933 & .933 \\
\bottomrule
\end{tabular*}

\end{table}

Acquisition costs and review scheduling for these pairs appear in
Appendix~\ref{app:timing}.

\FloatBarrier
\endgroup

\FloatBarrier
\Needspace{10\baselineskip}
\section{Teaching Process Analysis}

\subsection{Teaching-intervention taxonomy and dialogue turns}
\label{app:intents}
The Educator records intent labels when issuing each teaching tool call.
These labels describe its intended change in learner state.
A lesson may receive multiple labels; silence is a separate decision.
The codebook distinguishes task knowledge, reasoning, evidence acquisition,
diagnosis, calibration, and consolidation, following the emphasis on task,
process, and self-regulation in feedback research and active construction
during tutoring~\citep{hattie2007feedback,chi2001learning}.

\begin{table}[!htbp]
\caption{Intent definitions and tag frequencies for Qwen Fixed-Window SSL
($g=10$; \ReviewFamilies{} families, \ReviewRuns{} acquisition runs).
Opening lessons and follow-up turns use separate denominators
of \ReviewOpeningLessons{} and \ReviewFollowUps{}.}
\label{tab:intent-codebook}
\centering\small
\setlength{\tabcolsep}{4pt}
\renewcommand{\arraystretch}{1.12}
\begin{tabularx}{\linewidth}{@{}>{\raggedright\arraybackslash}p{.22\linewidth}>{\raggedright\arraybackslash}Xrr@{}}
\toprule
Intent & Operational target & Opening & Follow-up \\
\midrule
\ReviewIntentRows
\bottomrule
\end{tabularx}
\end{table}

Reasoning is the most common
opening intent; diagnosis and verification are more frequent in follow-ups.

\FloatBarrier
\Needspace{6\baselineskip}
\subsection{Review decisions and subsequent Student behavior}
\label{app:review-behavior}
We relate teaching and silence to the Student's next action and subsequent
Skill revision. The analysis covers \ReviewRuns{} \fixedwindow{} runs from
the evaluation manifest, using the latest complete attempt for each family
and repeat. They contain \ReviewTotal{} reviews, \ReviewOpeningLessons{}
opening lessons, and \ReviewFollowUps{} follow-ups. Regular ten-tool-call
reviews include \ReviewTeachIntervals{} teaching and \ReviewSilentIntervals{}
silent decisions; trajectory-end reviews include
\ReviewTerminalTeaching{} and \ReviewTerminalSilence{}, respectively.

\paragraph{Observation windows.}
Each window begins when the Student receives a lesson or a silent review
completes, and ends at the next review or round boundary. We use event order
and tool-call IDs to identify actions begun within the window. Initial
prompts delimit rounds; recovery messages remain within the current round.

Across \ReviewIntervalCount{} regular windows, we classify the first
subsequent tool call as environment, write/edit, read/search, or absent.
Trajectory-end reviews are analyzed separately:
\ReviewTerminalTeachingPreceded{}/\ReviewTerminalTeaching{} lessons precede
a tool call, while all \ReviewTerminalSilence{} silent reviews end the round.

\paragraph{Skill revision over ten tool calls.}
The revision analysis uses \ReviewPackageEligible{} windows containing
exactly ten calls begun after review, with no continuation of an earlier call.
We compare the Skill versions at the start and end of each window.
Both versions are observable in \ReviewPackageKnown{} windows; the other
\ReviewOtherWindows{} lack a comparable pair. Among these
\ReviewPackageKnown{} windows, the package changes in
\ReviewPackageTeachChanged{}/\ReviewPackageTeachN{} teaching windows and
\ReviewPackageSilentChanged{}/\ReviewPackageSilentN{} silent windows.
A sensitivity analysis relaxes the ten-call restriction and includes
\ReviewAllIntervalKnown{} windows with comparable versions.

\paragraph{Estimation.}
These observational comparisons measure associations with teaching:
review decisions respond to the Student's ongoing work. We standardize
contrasts by acquisition round, using rounds with both comparison groups
and weighting them by the eligible sample's round distribution.
Table~\ref{tab:review-behavior} reports the number of supported observations
with known outcomes ($n$). Confidence intervals resample the
\ReviewFamilies{} whole families, keeping each family's runs and reviews
together and retaining the round weights.
Pre-review comparisons use recorded tool composition, error flags, shell
exit status, and package state. Tool-error and no-error groups contain
\ReviewToolErrorWindows{} and \ReviewNoErrorWindows{} regular windows;
the observable/unobservable package comparison is supported in round one.

\paragraph{Writing and revision after teaching.}
Following teaching, the Student's next tool call is
\num[retain-explicit-plus=false]{\ReviewWriteAdjustedPP} percentage points
more likely to write or edit a file than following silence
(95\% CI: $\ReviewWriteCI$). Over the next ten tool calls, the Skill package
is \num[retain-explicit-plus=false]{\ReviewPackageAdjustedPP} points more
likely to change ($\ReviewPackageCI$), after standardizing by round.
A within-run check gives a mean package-change difference of
\ReviewWithinRunPackagePP{} points over \ReviewWithinRunPackageN{} runs
containing both decision types.
\begin{table}[H]
\centering\small
\caption{Round-standardized review-behavior contrasts for Qwen
\fixedwindow{} ($g=10$; \ReviewFamilies{} families, \ReviewRuns{} acquisition runs).
Counts give supported observation windows, and brackets
give cluster-bootstrap confidence intervals over task families.}
\label{tab:review-behavior}
\begin{tabularx}{\linewidth}{@{}Xrr@{}}
\toprule
Outcome and comparison & $n$ & Adjusted difference (pp) \\
\midrule
Teaching: observed tool error vs no error & 432 & \num{-0.4} [\num{-9.2}, \num{+8.1}] \\
Teaching: shell failure vs no failure & 413 & \num{-8.5} [\num{-20.8}, \num{+2.6}] \\
Teaching: observable vs unobservable package & 233 & \num{-31.5} [\num{-41.7}, \num{-22.8}] \\
Teaching: preceding package change vs none & 130 & \num{-10.2} [\num{-28.7}, \num{+12.4}] \\
Teaching: environment-dominant vs other work & 432 & \num{+1.5} [\num{-8.2}, \num{+10.1}] \\
Teaching: write-dominant vs other work & 432 & \num{+1.8} [\num{-21.4}, \num{+25.2}] \\
Teaching: read-dominant vs other work & 432 & \num{+4.6} [\num{-10.2}, \num{+19.4}] \\
Environment first: teaching vs silence & 432 & \num{+0.9} [\num{-11.5}, \num{+13.0}] \\
Write/edit first: teaching vs silence & 432 & \num{+23.9} [\num{+13.0}, \num{+36.1}] \\
Read/search first: teaching vs silence & 432 & \num{-17.2} [\num{-25.0}, \num{-8.7}] \\
No subsequent tool: teaching vs silence & 432 & \num{-7.7} [\num{-10.9}, \num{-4.5}] \\
Package change, ten tools: teaching vs silence & 112 & \num{+38.6} [\num{+10.7}, \num{+57.8}] \\
Package change, all intervals: teaching vs silence & 316 & \num{+60.1} [\num{+49.2}, \num{+70.6}] \\
\bottomrule
\end{tabularx}

\end{table}

\Needspace{9\baselineskip}
\subsection{Learning progress and review cost}

Each of the \ReviewRuns{} runs completes all \MainRounds{} acquisition rounds,
with practice continuing after anchor success. We link the resulting \ReviewTerminalCount{}
fresh-solver anchor evaluations to package snapshots. The unit of analysis
is a later round paired with its preceding anchor result, yielding
\ReviewTransitions{} transitions with comparable snapshots.

\begin{figure}[!hbp]
\centering
\includegraphics[width=\linewidth]{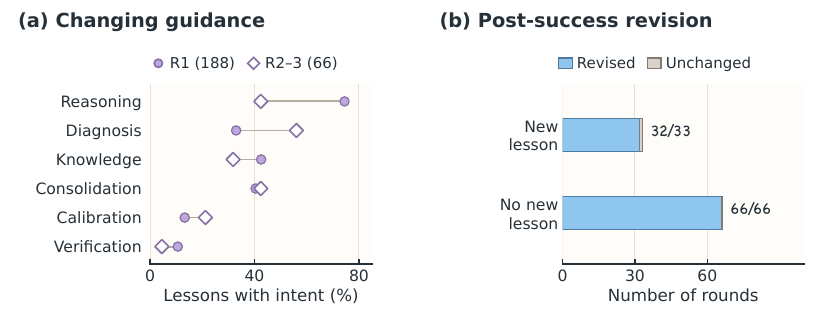}
\caption{\textbf{Teaching emphasis and development after anchor success.}
Qwen Fixed-Window SSL ($g=10$), \ReviewFamilies{} families and \ReviewRuns{} acquisition runs.
(a) Educator-reported intents in first-round versus later lessons; labels
may overlap. (b) Package revisions in rounds following a perfect anchor
score, grouped by whether a new lesson arrives.}
\label{fig:review-behavior-full}
\end{figure}

\paragraph{Teaching across rounds.}
Figure~\ref{fig:review-behavior}a uses regular-review lesson counts of
\RoundOneRegularTeach{}/\RoundOneRegularReviews{},
\RoundTwoRegularTeach{}/\RoundTwoRegularReviews{}, and
\RoundThreeRegularTeach{}/\RoundThreeRegularReviews{}. Including trajectory-end
reviews gives \StageOneTeach{}/\StageOneReviews{},
\StageTwoTeach{}/\StageTwoReviews{}, and \StageThreeTeach{}/\StageThreeReviews{}.
Utility on anchor tests by fresh solvers is \AnchorRoundOne{}, \AnchorRoundTwo{}, and
\AnchorRoundThree{}, with \StageOnePerfect{}/\ReviewRuns{},
\StageTwoPerfect{}/\ReviewRuns{}, and \StageThreePerfect{}/\ReviewRuns{} perfect
results. Figure~\ref{fig:review-behavior-full}a compares \StageFirstLessons{} first-round lessons with
\StageLaterLessons{} later lessons; reasoning tags occur in
\StageEarlyReasoningCount{} versus \StageLateReasoningCount{} and diagnosis tags
in \StageEarlyDiagnosisCount{} versus \StageLateDiagnosisCount{}.

\paragraph{Development after anchor success.}
Of the \ReviewTransitions{} later-round transitions, \PostPerfectTransitions{}
follow a perfect anchor (\PostPerfectSecond{} second rounds and
\PostPerfectThird{} third rounds), covering \PostPerfectRuns{} runs and all
\ReviewFamilies{} families. Packages change in \ProgressRevisedAfterPass{}/
\PostPerfectTransitions{}: \PostPerfectLessonChanged{}/\PostPerfectLessonTotal{}
rounds with a new lesson and \PostPerfectSilentChanged{}/\PostPerfectSilentTotal{}
without one. Those \PostPerfectLessonTotal{} rounds contain
\PostPerfectLessons{} lessons. Across all later rounds, the
\PostPerfectLaterSilent{} rounds without a new lesson all revise their package. Prior lessons and anchor feedback remain
available.

\paragraph{Review cost before follow-up.}
Silent reviews account for \ReviewSilentOpportunityShare{}\% of review
opportunities and \ReviewSilentOpeningShare{}\% of opening-review tokens.
Figure~\ref{fig:review-behavior}c includes cached input and excludes follow-up
dialogue. Each round's teaching and silent token totals are divided by its
total review count, so stacked heights give mean cost per review.
Round totals are \RoundOneTokenM{}M, \RoundTwoTokenM{}M, and
\RoundThreeTokenM{}M tokens; silent shares are \RoundOneSilentShare{}\%,
\RoundTwoSilentShare{}\%, and \RoundThreeSilentShare{}\%.
Average token use per review rises from \ReviewFirstCostK{}K to
\ReviewLastCostK{}K tokens per review across acquisition rounds, a
\ReviewCostGrowth{}-fold increase. Appendix~\ref{app:timing} reports
schedule-level costs including follow-up dialogue.

The corresponding teaching sessions and training-iteration evaluations are
released under \path{experiments/main/qwen/fixed-window-ssl/}; each run's \path{meta.json} lists
its session inventory.

\FloatBarrier
\Needspace{10\baselineskip}
\section{Review Scheduling and Cost Allocation}
\label{app:timing}

\subsection{Schedules and role-level cost}
\begin{figure}[htbp]
\centering
\includegraphics[width=\linewidth]{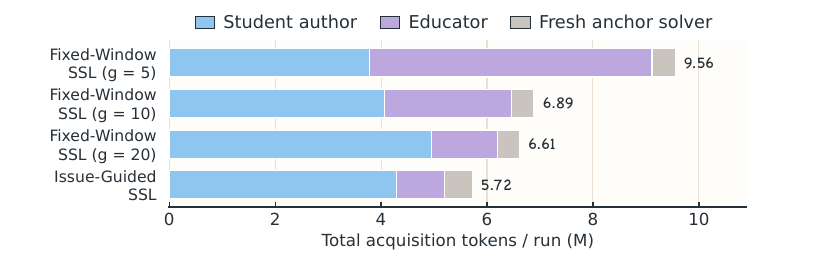}
\caption{\textbf{Acquisition-cost allocation across the four teaching schedules.}
Qwen, \MainFamilies{} families and \MainRuns{} acquisition runs per schedule.
Stacked bars show mean Student-author, Educator, and anchor-evaluation tokens
per run, with the total labeled for each schedule.}
\label{fig:schedule-role-costs}
\end{figure}

All four Qwen schedules use \MainFamilies{} families, \MainRounds{} acquisition rounds, and \MainRepeats{}
repeats (\MainRuns{} runs each). The ten-tool-call condition is the main Fixed-Window SSL experiment.
Table~\ref{tab:timing} reports utility, cost, and teaching activity (lessons
exclude follow-ups), with mean acquisition tokens split into Student-author,
Educator, and anchor-evaluation work (Figure~\ref{fig:schedule-role-costs}).
\begin{table}[!htbp]
\caption{Utility, acquisition cost, reviews, and opening lessons for the four
Qwen schedules on the main cohort (\MainFamilies{} families, \MainVariants{} cases;
\MainRepeats{} repeats and \MainRuns{} acquisition runs per schedule).}
\label{tab:timing}
\centering\small
\renewcommand{\arraystretch}{1.10}
\setlength{\tabcolsep}{4pt}
\begin{tabular*}{\linewidth}{@{\extracolsep{\fill}}l S[table-format=1.3] S[table-format=1.2] S[table-format=4.0] S[table-format=3.0]@{}}
\toprule
Teaching schedule & {Utility $\uparrow$} & {Tokens/run (M) $\downarrow$} & {Reviews} & {Lessons} \\
\midrule
Every 5 tool calls & \GFiveUtility & \GFiveTokens & \GFiveObservations & \GFiveLessons \\
Every 10 tool calls & \GTenUtility & \GTenTokens & \GTenObservations & \GTenLessons \\
Every 20 tool calls & \GTwentyUtility & \GTwentyTokens & \GTwentyObservations & \GTwentyLessons \\
\midrule
\shortstack[l]{Issue transitions\\(Issue-Guided SSL)} & \bfseries \TreeUtility & \bfseries \TreeTokens & \bfseries \TreeObservations & \bfseries \TreeLessons \\
\bottomrule
\end{tabular*}
\end{table}

\treedevelop{} reaches \TreeUtility{} utility versus \GTenUtility{} for
\fixedwindow{} at $g=10$, while reducing Educator tokens by
\TreeEducatorReduction{}\% and total acquisition tokens by
\TreeTotalCostReduction{}\% (\GTenTokens{}M to \TreeTokens{}M per run).
Teaching efficiency is \TreeEducatorEfficiencyRatio{} times that of
\fixedwindow{} at $g=10$.
The fixed $g=5$ and $g=20$ schedules reach \GFiveUtility{} and
\GTwentyUtility{}, respectively.

Fixed-Window SSL uses \ISTEducatorTotalM{}M Educator tokens across \GTenObservations{}
reviews; Issue-Guided SSL uses \TreeEducatorTotalM{}M across
\TreeObservations{}. The \ReviewReduction{}\% reduction in reviews combines
with a \ReviewTokenReduction{}\% reduction in tokens per review
(\ISTEducatorPerReviewK{}K to \TreeEducatorPerReviewK{}K), yielding a
\TreeEducatorReduction{}\% reduction in Educator tokens.
These per-review costs include follow-up dialogue. Issue-Guided SSL combines
issue-triggered review with local context and its follow-up policy; the
protocol settings are detailed in Appendix~\ref{app:protocol-loop}.

Table~\ref{tab:role-cost-balance} puts reductions in Educator tokens and
changed learner work on the same scale, in millions of tokens per run.
Positive net savings mean fewer acquisition tokens within the recorded scope.
Student tokens fall \KimiStudentReduction{}\% on Kimi but rise
\DeepSeekStudentIncrease{}\% on DeepSeek, where the additional Student work
exceeds the Educator saving. These outcomes distinguish the consistent
reduction in external supervision from the model-dependent change in
total acquisition cost.
\begin{table}[!htbp]
\caption{Role-level token changes and net acquisition savings for
Issue-Guided SSL relative to Fixed-Window SSL, in millions of tokens per run.
Qwen uses the \MainFamilies{} main-cohort families (\MainRepeats{} repeats);
DeepSeek and Kimi use the \CrossFamilies{} shared families (one repeat).
DeepSeek and Kimi rows cover the persistent author and Educator sessions.}
\label{tab:role-cost-balance}
\centering\small
\setlength{\tabcolsep}{4pt}
\begin{tabular*}{\linewidth}{@{\extracolsep{\fill}}lrrrrr@{}}
\toprule
Pair & Runs & \shortstack{Educator\\saved} & \shortstack{Author\\added} & \shortstack{Anchor\\added} & \shortstack{Net\\saved} \\
\midrule
Qwen & 66 & 1.479 & 0.218 & 0.097 & 1.164 \\
DeepSeek & 10 & 4.175 & 5.276 & -- & $-$1.101 \\
Kimi & 10 & 3.250 & $-$0.870 & -- & 4.120 \\
\bottomrule
\end{tabular*}

\end{table}

On DeepSeek, \DSStudentAddedM{}M additional Student tokens arise mainly from
more calls (\DSStudentCallsIST{} to \DSStudentCallsTree{},
+\DSStudentCallIncreasePercent{}\%), concentrated in the first two rounds.
\texttt{sqlite-db-truncate} adds \DSSqliteAddedM{}M total acquisition tokens;
the other families save \DSOtherFamiliesSavedM{}M, leaving a net increase of
\DSNetAddedM{}M. The \DSEducatorSavedM{}M Educator saving combines
fewer reviews (\DeepSeekISTWakeups{} to \DeepSeekTreeReviews{}) with smaller
ones. Fixed-Window SSL has \DSGiantCalls{} Educator calls re-reading over
500K cached tokens (maximum \DSGiantCallMaxK{}K); its top 15\% of calls
consume \DSGiantCallTokenShare{}\% of Educator tokens. Issue-Guided SSL has
no calls above 500K cached tokens.
\begin{table}[!htbp]
\caption{Per-run acquisition tokens by session role for Fixed-Window SSL and
Issue-Guided SSL for DeepSeek and Kimi on the \CrossFamilies{}-family lite subset
(single repeat), covering
the persistent author (Student) and Educator sessions.}
\label{tab:cross-model-tokens}
\centering\small
\setlength{\tabcolsep}{5pt}
\begin{tabular*}{\linewidth}{@{\extracolsep{\fill}}ll rrr@{}}
\toprule
Pair & Schedule & \shortstack{Student\\(M/run)} & \shortstack{Educator\\(M/run)} & \shortstack{Total\\(M/run)} \\
\midrule
DeepSeek & \fixedwindow & 13.46 & 6.55 & 20.01 \\
 & \treedevelop & 18.74 & 2.37 & 21.11 \\
Kimi & \fixedwindow & 5.41 & 4.20 & 9.61 \\
 & \treedevelop & 4.54 & 0.95 & 5.49 \\
\bottomrule
\end{tabular*}

\end{table}

\FloatBarrier
\Needspace{6\baselineskip}
\subsection{Issue lifecycles}
Table~\ref{tab:issue-lifecycle} distinguishes persistent nodes from repeated
transition events. A closure ends a review episode and need not be the issue's
last appearance. Most closures record Educator silence; verification and
bounded-scope dispositions bypass another review in \IssueBypassClosures{} events. All \IssueReopenEvents{} reopen
events lead to another lesson. The Flask sequence in
Appendix~\ref{app:flask-chain} illustrates this reopening behavior at the event level.
\begin{table}[H]
\caption{Qwen Issue-Guided SSL lifecycle accounting across \MainFamilies{} families
and \MainRuns{} acquisition runs, listing persistent issues, reopen events, closure dispositions, and issue
status at run end.}
\label{tab:issue-lifecycle}
\centering\small
\begin{minipage}[t]{.57\linewidth}
\vspace{0pt}
\begin{tabularx}{\linewidth}{@{}>{\raggedright\arraybackslash}Xr@{}}
\toprule
Persistent issues & Count or summary \\
\midrule
\IssueLifecycleNodeRows
\bottomrule
\end{tabularx}
\end{minipage}\hfill
\begin{minipage}[t]{.39\linewidth}
\vspace{0pt}
\begin{tabularx}{\linewidth}{@{}>{\raggedright\arraybackslash}Xr@{}}
\toprule
Review closures & Count \\
\midrule
\IssueLifecycleClosureRows
\bottomrule
\end{tabularx}
\end{minipage}
\end{table}

In an API-scheduling no-lesson example,
prior anchor feedback shows the solver searching the wrong script location
and reimplementing the scheduler. The Student adds execution and Quick Start
guidance to \texttt{SKILL.md}; review closes as \texttt{teacher-silence}.

\FloatBarrier
\Needspace{10\baselineskip}
\section{Case Evidence and Frozen-Package Use}
\label{app:chains}

\subsection{Testing and adapting Educator guidance}
The HTML-finance case comes from Qwen \fixedwindow{}, repeat 1, in the
main \MainFamilies{}-family cohort. It follows a teaching proposal through
Student execution, package revision, and fresh use
(Table~\ref{tab:case-test}).

\paragraph{From a proposal to a checked procedure.}
The Educator recommends deriving reported financial values from other entries
and distinguishing agreement, disagreement, and the absence of an applicable
rule. It also suggests generic arithmetic search. The Student implements
and runs this strategy on the anchor. The resulting verifier flags
\CaseInitialFlags{} metrics and selects line 140, although the incorrect
entry is the gross margin at line \CaseCorrectLine{}.
Inspecting the output, the Student identifies missing financial derivations
and replaces generic arithmetic with relations appropriate to each metric,
including reverse derivations. Running the revised verifier on the same
report leaves only line \CaseCorrectLine{} flagged: the reported gross margin
is 60.8\%, while $520.4/842.0\times100$ rounds to 61.8\%.
The contrast records two executions during acquisition.

\paragraph{Retained knowledge and fresh use.}
The final \texttt{SKILL.md} retains three outcomes---confirmed, contradicted,
and unchecked---and instructs solvers to use both forward and reverse
financial relations. It bundles the revised \texttt{verify.py}.
On held-out \texttt{v14}, a fresh solver reads both resources and checks
relations in a cash-flow statement. Operating, investing, and financing
subtotals sum to 8,000, whereas the reported net increase is 6,270.
The closing cash balance independently agrees with 8,000. The solver submits
\texttt{4206,6270}, identifying the erroneous line and its reported value,
and passes \CaseFreshPassed{}/\CaseFreshTotal{} checks.

\paragraph{Source records.}
The released run is \path{experiments/main/qwen/fixed-window-ssl/r1/html-finance-verify/}.
Within it, \path{skill-formation/student-author.jsonl} contains the lesson
and two executions (records 20 and 42--44); \path{heldout/v14.jsonl}
contains package reads, accounting checks, and submission (records 4, 10,
12, and 14). Scores for all six held-out cases are in \path{meta.json}.
Record numbers refer to the JSONL \texttt{i} field.

\subsection{Flask issue reopening: event-level evidence}
\label{app:flask-chain}
In a Qwen Issue-Guided SSL Flask run, \texttt{issue-1} receives a
lesson, revision, and no-lesson closure during the first round. Fresh-solver anchor
feedback with 6/9 checks and missing-table errors reopens it in the next round. The renewed
lesson identifies an import-bound database path: a changed working directory
can redirect the service to the wrong database. The Student reproduces
the mismatch, resolves paths at request time, and reports \FlaskLocalChecks{}/\FlaskLocalChecks{} local
checks before verification-backed closure. Separately,
fresh-solver anchor utility is \FlaskAnchorOne{}, \FlaskAnchorTwo{}, and
\FlaskAnchorThree{} across rounds.
The released run is \path{experiments/main/qwen/issue-guided-ssl/r3/flask/}.
Its session logs are restored with the command in Appendix~\ref{app:implementation}.
In \path{skill-formation/student-author.jsonl}, records 50 and 52 show
the anchor feedback and renewed guidance; records 80--82 show verification
and the revision addressing the issue.

\FloatBarrier
\Needspace{7\baselineskip}
\section{Limitations}
\label{app:limitations}
Our study evaluates complete Skill-learning protocols under a shared
single-anchor setting. Further controlled comparisons could isolate how
much the Educator, review timing, and issue structure each contribute to
transfer performance and teaching cost.

\end{document}